\documentclass[conference]{IEEEtran}
\IEEEoverridecommandlockouts
\usepackage{cite}
\usepackage{amsmath,amssymb,amsfonts}
\usepackage{graphicx}
\usepackage{textcomp}
\usepackage{algorithm, algpseudocode}
\usepackage{subcaption}
\usepackage{xcolor, soul}
\usepackage{float}% http://ctan.org/pkg/float
\usepackage{url}

\sethlcolor{yellow}
\def\BibTeX{{\rm B\kern-.05em{\sc i\kern-.025em b}\kern-.08em
    T\kern-.1667em\lower.7ex\hbox{E}\kern-.125emX}}
    
\begin{document}

\title{Joint Level Generation and Translation Using Gameplay Videos\\}

\author{\IEEEauthorblockN{1\textsuperscript{st} Negar Mirgati}
\IEEEauthorblockA{\textit{dept. of Computing Science, Amii} \\
\textit{University of Alberta}\\
Edmonton, Canada \\
mirgati@ualberta.ca}
\and
\IEEEauthorblockN{2\textsuperscript{nd} Matthew Guzdial}
\IEEEauthorblockA{\textit{dept. of Computing Science, Amii} \\
\textit{University of Alberta}\\
Edmonton, Canada \\
guzdial@ualberta.ca}
}
\IEEEoverridecommandlockouts
\IEEEpubid{\makebox[\columnwidth]{979-8-3503-2277-4/23/\$31.00~\copyright2023 IEEE \hfill} 
\hspace{\columnsep}\makebox[\columnwidth]{ }}
\maketitle
\IEEEpubidadjcol

\begin{abstract}
Procedural Content Generation via Machine Learning (PCGML) faces a significant hurdle that sets it apart from other fields, such as image or text generation, which is limited annotated data. Many existing methods for procedural level generation via machine learning require a secondary representation besides level images. However, the current methods for obtaining such representations are laborious and time-consuming, which contributes to this problem.
 In this work, we aim to address this problem by utilizing gameplay videos of two human-annotated games to develop a novel multi-tail framework that learns to perform simultaneous level translation and generation. The translation tail of our framework can convert gameplay video frames to an equivalent secondary representation, while its generation tail can produce novel level segments. Evaluation results and comparisons between our framework and baselines suggest that combining the level generation and translation tasks can lead to an overall improved performance regarding both tasks. This represents a possible solution to limited annotated level data, and we demonstrate the potential for future versions to generalize to unseen games.
\end{abstract}


\section{Introduction}
Procedural Content Generation via Machine Learning (PCGML) is the practice of generating new game content by applying various machine learning (ML) techniques to existing game data or environments \cite{summerville2018procedural}. Although there has been considerable advancement in this area, one of the fundamental challenges PCGML researchers face is the limited availability of annotated data. Contrary to other ML tasks, such as image generation, PCGML methods typically need more information than the raw pixel representation of games. This additional requirement arises from the fact that game data, such as levels, have special structural features and functional rules, which are hard to capture solely from images. For instance, consider the lava blocks in Super Mario Bros. castle levels. It is hard to determine the behavioral properties (e.g., that it will kill the player) of these blocks from their visual appearance alone. Consequently, we require additional annotation to generate new game content that closely follows the same structure and abides by the constraints of the original data \cite{Jadhav_Guzdial_2021}.

One of the most commonly used alternative approaches to represent game levels is by translating tiles of pixels to text characters, where each character stands in for an object in the level. The conventional procedure of generating string representations of game levels usually involves multiple iterations of image processing via tools such as OpenCV \cite{opencv_library}, followed by manually editing the results, which is quite labor-intensive \cite{Guzdial_Riedl_2021}. In this regard, The Video Game Level Corpus (VGLC) \cite{summerville2016vglc} has immensely assisted PCGML researchers by providing public access to string representations of 428 levels from 12 different platformer-based 2D games. Though this is an impressive range of coverage in the area of games, it is still orders of magnitude smaller than numerous widely-used datasets (e.g., ImageNet \cite{deng2009imagenet}, and CelebA \cite{liu2015deep}) that are used in image generation. If we could access sufficient data, we may be able to achieve the success and generalizability of modern large language models and image generators.

A less explored yet promising secondary source of data for PCGML is gameplay video. The active community of video game enthusiasts has produced countless gameplay videos across genres, including platformers, which are currently the target of a large proportion of PCGML research. Despite the existence of significant video data, few approaches adapt gameplay video for game level generation. The existing methods either focus on a single game \cite{Guzdial_Riedl_2021} or focus on translating the data to another representation, instead of level design \cite{NEURIPS2021_2bcab9d9}.

%Guzdial and Riedl \cite{Guzdial_Riedl_2021} proposed an unsupervised approach to training a probabilistic model that utilizes gameplay video to generate new levels. Although this model successfully captures the style of the original game of training data, \hl{it does not generalize over new unseen games}. Furthermore, Smirnov et al. \cite{NEURIPS2021_2bcab9d9} introduced a novel unsupervised framework that is able to extract recurring sprites from gameplay videos. \hl {However, existing PCGML methods cannot directly use this form of data to generate new levels}.%

Motivated by the problems discussed above, we propose a novel machine learning framework for simultaneous video to level translation and generation. Our final model translates gameplay video into a tile-based string representation and simultaneously generates new level segments. We accomplish this with a modified VAE-GAN architecture, with each tail handling one task. We train our approach with YouTube videos and the VGLC level corpus but demonstrate the ability to generalize over content outside the VGLC representation.

%Paragraph 1 %
% #TODO: What is the broader field? %
% (1) start with PCG/PCGML definition
% (2) data problems
% (3) data processing problems 
% (4) What if we could automatically go from an unannotated representation to an annotated one during the training process (or something)

% Paragraph 2 %
% (1) The original VGLC paper "Traditional" way with the VGLC and OpenCV or similar
% (2) Marionette
% (3) Game Level Generation from Gameplay Videos

%Paragraph 3 % - high level overview
% Very high level pitch of the work, rely on types of approaches/phrases. We employed a unique VAE-GAN setup to process video data into annotated tile-based level data, etc. 

%Paragraph 3-4 %
% state contributions 
Our contributions can be summarized as follows:
\begin{enumerate}
    \item We propose the problem of simultaneous level translation and generation for the first time.
    \item We develop a novel architecture that aims to address this new problem.
    \item We demonstrate evidence showing that our novel architecture outperforms relevant baselines for this problem.
\end{enumerate}

%Paragraph 5 (optional) %
The rest of this paper is organized as follows: In sections \ref{sec:rel_work} and \ref{sec:background}, we briefly overview related work. Sections \ref{sec:approach}, \ref{sec:eval}, and \ref{sec:results} explain our approach, evaluation methods, and results. Section \ref{sec:future} discusses the limitations and some future directions of work. Finally, section \ref{sec:conc} concludes this paper.

\section{Related Work}\label{sec:rel_work}
%(Maybe have a bit of text up here on why these two areas are being covered)%
We divide this section into two parts based on the two tasks of our problem. First, we look into works that focus on the task of level translation. Second, we review the most related approaches to PCGML level generation.

\subsection{Level Translation}
Level translation is the process of converting the raw pixel representation (image) of a game level into a secondary representation that is useful for downstream tasks. As an example, Summerville et al. \cite{summerville2016vglc} created the VGLC corpus that contains level images and their corresponding string representations. In the level images, each $16 \times 16$ pixel tile specifies a game object (e.g., enemy, door, coin), and the VGLC representation converts each tile to a character. As a result, it provides a sequence of characters (i.e., string) for each level image. It is most likely that the creators of the VGLC corpus utilized a non-ML approach involving the OpenCV library \cite{opencv_library} to perform the tile-to-character translation. Although we use the VGLC representation in this work, we propose employing an ML model to translate from pixels to the character representation automatically.

Chen et al. \cite{Chen_Sydora_Burega_Mahajan_Abdullah_Gallivan_Guzdial_2020} leveraged a Convolutional Neural Network (CNN) to translate $32\times32$ level segments to their corresponding string representation. They iteratively performed this process to convert an entire image to its equivalent string representation. Instead, we employ a Variational Autoencoder (VAE) to translate each video frame to its equivalent string representation at once.

Smirnov et al. \cite{NEURIPS2021_2bcab9d9} introduced a novel framework that can extract recurring sprites from gameplay videos. Their translated representation was not intended for level design tasks, unlike our own, and instead focuses on game reconstruction and scene understanding.%This type of representation is most suitable for editing or analysis purposes, whereas our tile-based string representation is most suitable for generation tasks.

Snodgrass and Onta\~n\'on \cite{snodgrass2016approach} introduced an approach to transform levels from one domain to levels of another by translating the input domain's tiles to the target domain. In comparison, we design a framework that learns to translate level images to the equivalent string representation in the same domain. However, we do train our approach on multiple domains, meaning we are capable of generating similarly `blended' levels. 

\subsection{PCGML Level Generation}
%There's a whole bunch of work on generating levels with PCGML (Cite the main PCGML journal paper). 
%Gameplay video Guzdial and Riedl
%Level Blending
%Mrunal's tile embedding
%(how you different)%

Researchers have suggested various methods for generating levels through the use of PCGML \cite{summerville2018procedural}. Relevant to our work, Guzdial and Riedl \cite{Guzdial_Riedl_2021} proposed an approach utilizing gameplay video to generate new levels. While their model successfully captures the style of the original game of training data, it does not generalize to unseen games since it requires authoring new code and a spritesheet for any new game. 

A number of existing works focus on generating blended levels. For instance, Sarkar, Yang, and Cooper \cite{sarkar19controllable} trained a VAE on two games to generate new blended level segments. Sarkar et al. \cite{sarkar2020exploring} then built upon the approach proposed in \cite{sarkar19controllable} and expanded the input domain to six games. While our approach also trains on multiple domains and generates blended levels, this is not the primary focus of our work. 

Jadhav and Guzdial \cite{Jadhav_Guzdial_2021}, and Khameneh and Guzdial \cite{khameneh2020embedding} focused on learning new representations that can be used for downstream tasks. These methods can be used as alternatives for dealing with the problem of insufficient training data in PCGML. We differ from these two approaches in the use of video data and in our unique VAE-GAN-based architecture that learns to translate video frames and generate new level segments simultaneously.

\section{Background}\label{sec:background}
In this section, we will briefly cover the deep learning architectures on which our VAE-GAN-based framework relies.

\subsection{GAN}
A Generative Adversarial Network (GAN) \cite{goodfellow2020generative} is a deep neural network that consists of a generator and a discriminator that compete against each other. The generator's goal is to output content that is hard to distinguish from the original content by the discriminator. On the other hand, the discriminator aims to strengthen its ability to differentiate the generator's outputs from the real data. The generator and discriminator are trained in an alternating manner until convergence. In our framework, we leverage the Wasserstein GAN (WGAN) \cite{arjovsky2017wasserstein}, which is a variant of the vanilla GAN that uses the Earth-Mover's distance (EM) instead of the Jensen-Shannon divergence.

\subsection{VAE}

Variational Autoencoder (VAE) \cite{Kingma2014} is a probabilistic generative model that consists of an encoder and a decoder. The encoder receives $X$ as input and outputs a latent distribution with mean $\mu$ and variance $\sigma$. The decoder then samples from this distribution to output $\tilde{X}$, a reconstruction of $X$. The model is trained with the objective of minimizing reconstruction loss as well as maintaining a Gaussian structure in the latent space.

\subsection{VAE-GAN}
A VAE-GAN \cite{larsen2016autoencoding} is a deep learning architecture that combines a VAE and a GAN by unifying the decoder of the VAE with the generator of the GAN. Therefore, this model is capable of encoding, generating, and differentiating between data samples. Larsen et al. \cite{larsen2016autoencoding} also provided evidence that the VAE-GAN architecture outperforms the VAE and the GAN regarding the quality of generated images, providing motivation to apply it to our problem. To the best of our knowledge, this architecture has not been previously applied to a level translation task.

\begin{figure*}
\centering
  \includegraphics[width=1\textwidth]{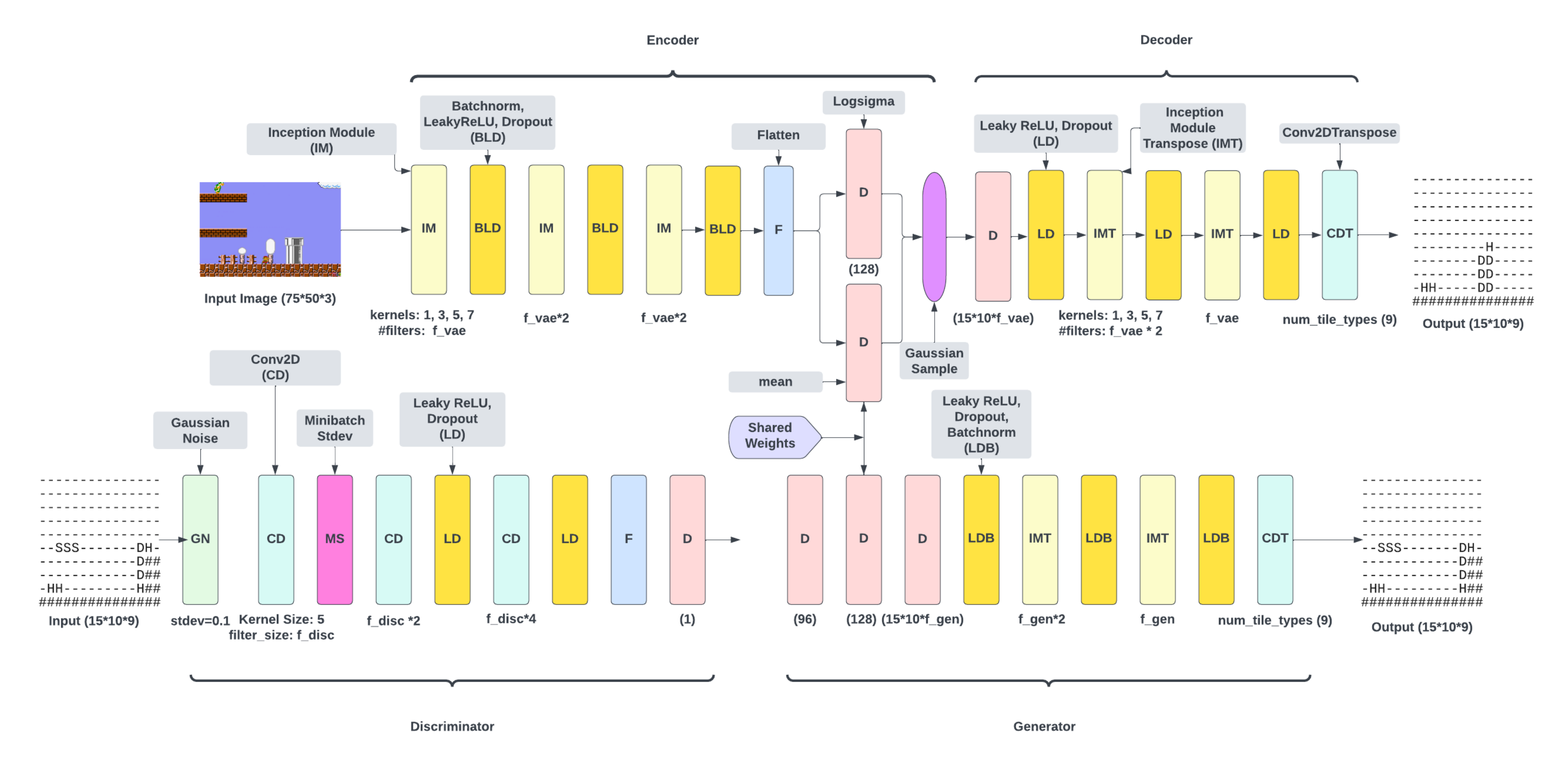}
  \caption{Our framework's architecture.}\label{fig:architecture}
\end{figure*}

\section{System Overview}\label{sec:approach}
 In this paper, we seek to train a single model on gameplay video capable of both level generation and translation. This section describes the procedure of implementing and training such a model in three parts. The first part outlines the primary steps we undertook to collect our dataset. The second part provides a detailed explanation of our model architecture. Lastly, the third part covers the training process of our framework.

\subsection{Dataset}
We required a dataset that matched gameplay video frames to level structure for our problem. To create this dataset, we undertook the following steps. First, we obtained high-quality gameplay videos of Super Mario Bros. and Kid Icarus from Youtube. Second, we used a video parser script \footnote{https://github.com/mguzdial3/VideoParser} to parse each video into a series of frames. We used $2$ frames per second (FPS) for Super Mario Bros. and $1$ for Kid Icarus. This helped keep multiple frames with the exact same level structures out of our dataset. Third, we resized all frames so that the tiles size was $16\times16$ pixels, which is the same as the tile size of level images in the VGLC. Fourth, we used the VGLC's level images to match the gameplay frames to the VGLC tile-based string representation. We employed OpenCV's template matching algorithm to look for the closest match of each frame in its corresponding level image. Using the location of the closest match, we paired the frame data with the appropriate string representation.

As the VGLC string representations differ between Super Mario Bros. and Kid Icarus, following the prior work \cite{sarkar2020exploring, Jadhav_Guzdial_2021}, we unified the string representations by using a common set of 9 tile types: 
\{(\#: solid, ground), 
(-: empty, background), 
(D: pipe, door),
 (H: enemy, harmful),
 (M: moving),
 (T: solid-top), 
(B: block), (S: breakable),
 (O: collectible)\}.
Fifth, we chose $10\times15$ as the string representation size and cropped our frame images to match the corresponding translations. We chose this string size since this was the smallest common window size that would work with the VGLC games. Finally, we downsized all frame images to $75x50$  since this frame size retained enough pixel information for the tiles to be recognizable and allows for odd-sized filters in the convolutional layers of our framework. We split our final dataset into training and test sets by setting aside Level $6$ of Kid Icarus and Worlds $7$ and $8$ of Super Mario Bros. as the test data. In addition, we upsample the Super Mario Bros. level segments in our training set to have an equal number of samples from both games. Our final dataset consists of $3956$ (frame, string translation) pairs, half for each game, in the training set and $360$ pairs in the test set.

\subsection{Model Architecture}
Our architecture is based on a typical VAE-GAN architecture introduced by \cite{larsen2016autoencoding}, with the VAE and GAN components. The VAE is primarily responsible for translating frames into a tile-based string representation. Meanwhile, the GAN is in charge of generating new level segments. Figure \ref{fig:architecture} demonstrates our model architecture. The defining characteristic of our VAE-GAN is the split of the Generator and the Decoder and the shared weights between the Encoder and Generator, which we included to help the model learn a latent space that would benefit both tasks.

Our model's VAE has two parts - an encoder and a decoder. The encoder receives frame images of size $75\times50\times3$ as input, where $3$ stands for the number of RGB channels. Our encoder is made up of consecutive Inception modules \cite{szegedy2015going}, followed by batchnorm \cite{ioffe2015batch}, leaky ReLU, and dropout layers. After all these layers, our encoder has a shared dense layer, and a variance dense layer, both of which have $d$ neurons, where $d$ represents the size of the latent space. Finally, the last layer of our encoder uses the mean $\mu$ and the variance $\sigma$ from the previous layers to sample a $d$-dimensional vector from the latent space. Our decoder receives the $d$-dimensional vector as its input from the encoder. It then passes it through consecutive sets of transposed Inception modules, leaky ReLU, and dropout, followed by a Convolutional Transpose layer with nine filters. The output of our decoder, which has size $10\times15\times9$, is a reconstruction of the input in the tile-based string representation. 

In our framework, we have made a modification to the original VAE-GAN architecture. Therefore, our GAN consists of a separate generator and a discriminator. To elaborate further, instead of having a single unified decoder/generator, we use separate decoder and generator networks to avoid overcomplicating the job of the single decoder/generator network. However, we allow the VAE and GAN components to share information by sharing the weights of the encoder's mean layer with the generator's second dense layer. Apart from the shared layer and an initial additional dense layer to facilitate the weight sharing, our generator has a very similar architecture to the decoder and the same input and output sizes.

Finally, the discriminator network consists of a Gaussian noise layer, a convolutional layer, a minibatch standard deviation layer \cite{karras2018progressive}, and two consecutive convolutional layers followed by leaky ReLU and dropout layers, and a final single-neuron dense layer. Notably, the last dense layer doesn't have any activation functions because our GAN follows the properties of the WGAN variation. The input size of this network is $10\times15\times9$, and it outputs a 1-dimensional value representing the model's evaluation score for the input.

\subsection{Model Training} 
Our VAE-GAN architecture has a unique optimization process. We train our framework as a whole, meaning that our VAE and GAN components are trained in a joint manner. In each epoch, we go through three primary stages. Firstly, we train our encoder and decoder to learn to perform the frame translation task. Secondly, following the training procedure of WGAN with gradient penalty \cite{gulrajani2017improved}, we train our discriminator towards optimality by training it for ten steps. Finally, in the last stage, we train our generator for a single step. 

We refer to algorithm \ref{alg:training} (adapted from \cite{larsen2016autoencoding}) for an overview of the training process.

Our framework is implemented using the Keras library \cite{chollet2015keras}. The training hyperparameters of our models are as follows: alpha coefficient of all leaky ReLU layers: $0.1473$, discriminator number of CNN filters (f\_disc): $8$, generator number of CNN filters (f\_gen): $2$, VAE number of CNN filters (f\_vae): $2$, discriminator dropout rate: $0.4684$, generator dropout rate: $0.2400$, VAE dropout rate: $0.2426$, latent space size: $128$, GAN learning rate: $1e-4$. We obtained these parameters by tuning the GAN and the VAE. The tuning process is covered in more detail in the supplementary materials.
%{'alpha rate': 0.14732279576819815,
% 'batch_size': 8,
 %'depth_disc': 8,
 %'depth_gen': 2,
 %'depth_vae': 2,
 %'dropout rate disc': 0.46847727119876675,
 %'dropout rate gen': 0.24003871439556243,
 %'dropout rate vae': 0.24261279625229554,
 %'latent_dim': 128,
 %'learning_rate': 0.0001}

\begin{algorithm} \label{alg:training_alg}
 \caption{Our VAE-GAN with gradient penalty. We use default values of $n_{disc}=10$, $\lambda = 10$} \label{alg:training}
 \begin{algorithmic}
 \State Initialize network parameters $\leftarrow{}$ $\theta_{Enc}$, $\theta_{Dec}$, $\theta_{Gen}$, $\theta_{Disc}$ 
\Repeat
  \State // Stage 1: VAE Training
 \State $ (X, Y) \leftarrow{}$ random mini-batch from dataset
 \State $Z \leftarrow{} Enc(X)$
 \State $\mathcal{L}\textsubscript{prior}\leftarrow{} D_{KL}(q(Z|X)||p(Z))$
 \State $\Tilde{X} \leftarrow{} Dec(Z)$
 \State $\mathcal{L\textsubscript{reconstruction}}$ $\leftarrow 
 Categorical\_Cross\_Entropy(\Tilde{X}, Y)$
 \State $\theta_{Enc}  \stackrel{+}\leftarrow - \triangledown_{\theta_{Enc}}({\mathcal{L}\textsubscript{prior} + \mathcal{L}\textsubscript{reconstruction}})$

 \State $\theta_{Dec}  \stackrel{+}\leftarrow - \triangledown_{\theta_{Dec}}({\mathcal{L}\textsubscript{prior} + \mathcal{L}\textsubscript{reconstruction}})$
 \\ \State // Stage 2: Discriminator Training
 \For{$t = 1, ..., n_{disc}$}
        \State $Z_{p} \leftarrow$ samples from prior $N (0, I)$
        \State $g_{fake} \leftarrow Gen(Z_{p})$
        \State $d_{fake} \leftarrow Disc(g_{fake})$
        \State $d_{real} \leftarrow Disc(Y)$
        \State $gp \leftarrow gradient\_penalty(Y, g_{fake})$
        \State $\mathcal{L}\textsubscript{Disc} \leftarrow 
        mean(d_{fake}) - mean(d_{real}) + \lambda * gp $
        \State $\theta_{Disc}  \stackrel{+}\leftarrow - \triangledown_{\theta_{Disc}}\mathcal{L}\textsubscript{Disc}$
\EndFor
 \\ \State // Stage 3: Generator Training
 \State $Z_{p} \leftarrow$ samples from prior $N (0, I)$ 
 \State $g_{fake} \leftarrow Gen(Z_{p})$
 \State $d_{fake} \leftarrow Disc(g_{fake}) $
  \State $\mathcal{L}\textsubscript{Gen} \leftarrow - mean(d_{fake})$
\State $\theta_{Gen}  \stackrel{+}\leftarrow - \triangledown_{\theta_{Gen}}\mathcal{L}\textsubscript{Gen}$
\Until deadline
 \end{algorithmic} 
 \end{algorithm}

\begin{table*}[!h]
\centering
\begin{tabular}{|c | c | c | c | c| c| c| c |} 
 \hline
 Model & Training Accuracy & Test Accuracy & Training E-distance & Test E-distance & Playability (SMB) & Playability (KI) \\ 
 \hline
 Our VAE-GAN & \textbf{0.94} & \textbf{0.88} & \textbf{0.31} & 
 \textbf{0.39} & 0.86 & 0.54 \\ 
 Original VAE-GAN & 0.90 & 0.85 & 0.37 & 0.44 & 0.85 & \textbf{0.60} \\
 GAN & - & - & 0.64 & 0.83 & 0.85 & 0.54 \\
 VAE & \textbf{0.94} & 0.87 & 1.69 & 0.84 & \textbf{0.94} & 0.58 \\
VAE-GAN\_TEXT & 0.92 & 0.87 & 0.66 & 1.81 & 0.85 & 0.54 \\
VAE\_TEXT & 0.92 & \textbf{0.88} & 2.53 & 2.55 & 0.85 & 0.47 \\  
 \hline
\end{tabular}
\caption{Results of generation and translation evaluation metrics for our model and all baselines.} \label{table:table_1}
\end{table*}

\begin{figure*}
     \centering
     \begin{subfigure}[b]{0.23\textwidth}
         \centering
         \includegraphics[width=\textwidth]{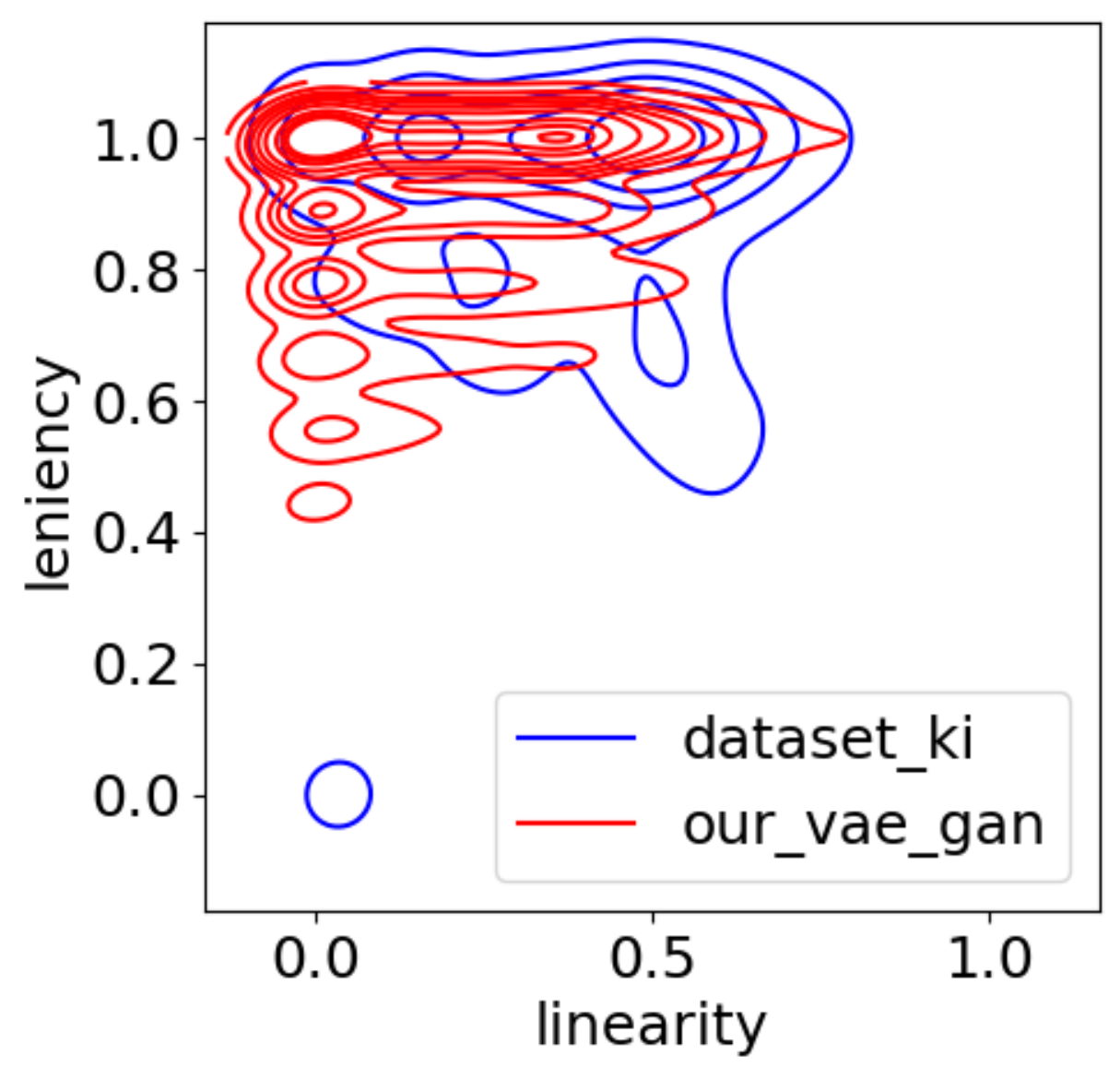}
         \caption{}
         \label{fig:ki_our_vae_gan}
     \end{subfigure}
     \begin{subfigure}[b]{0.23\textwidth}
         \centering
         \includegraphics[width=\textwidth]{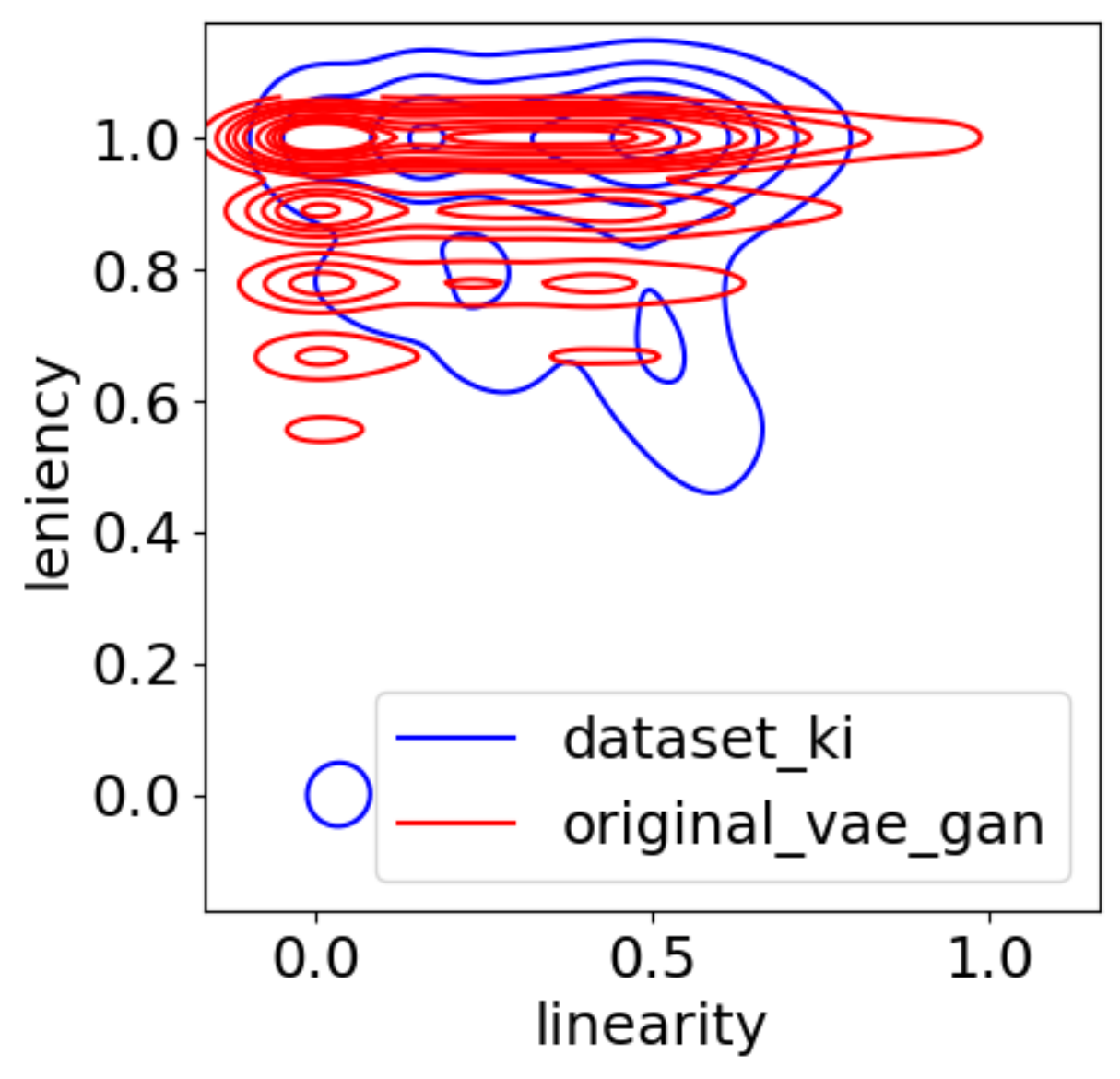}
         \caption{}
         \label{fig:ki_original_vae_gan}
     \end{subfigure}
     \begin{subfigure}[b]{0.23\textwidth}
         \centering
         \includegraphics[width=\textwidth]{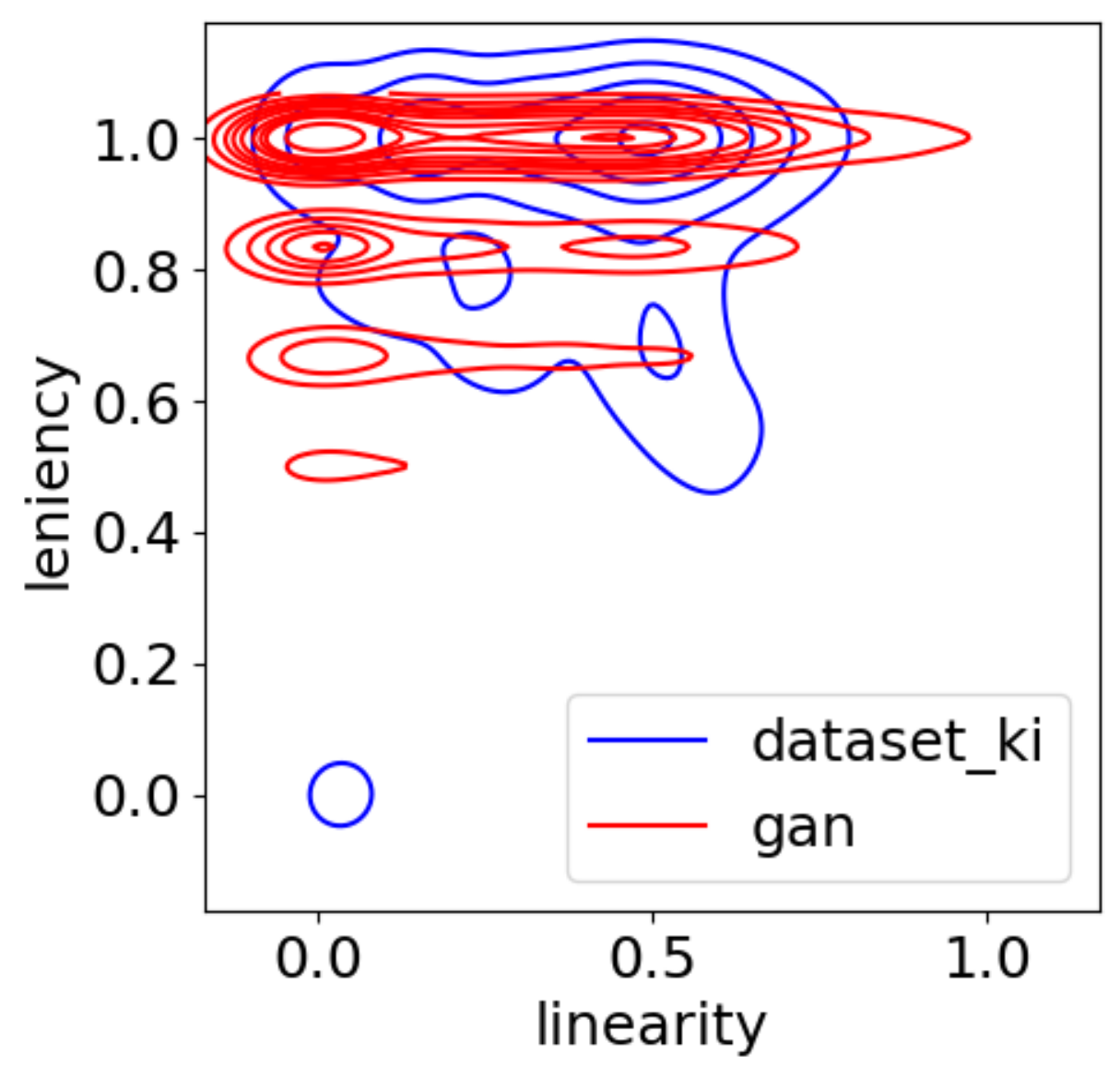}
         \caption{}
         \label{fig:ki_gan}
     \end{subfigure}
     \begin{subfigure}[b]{0.23\textwidth}
         \centering
         \includegraphics[width=\textwidth]{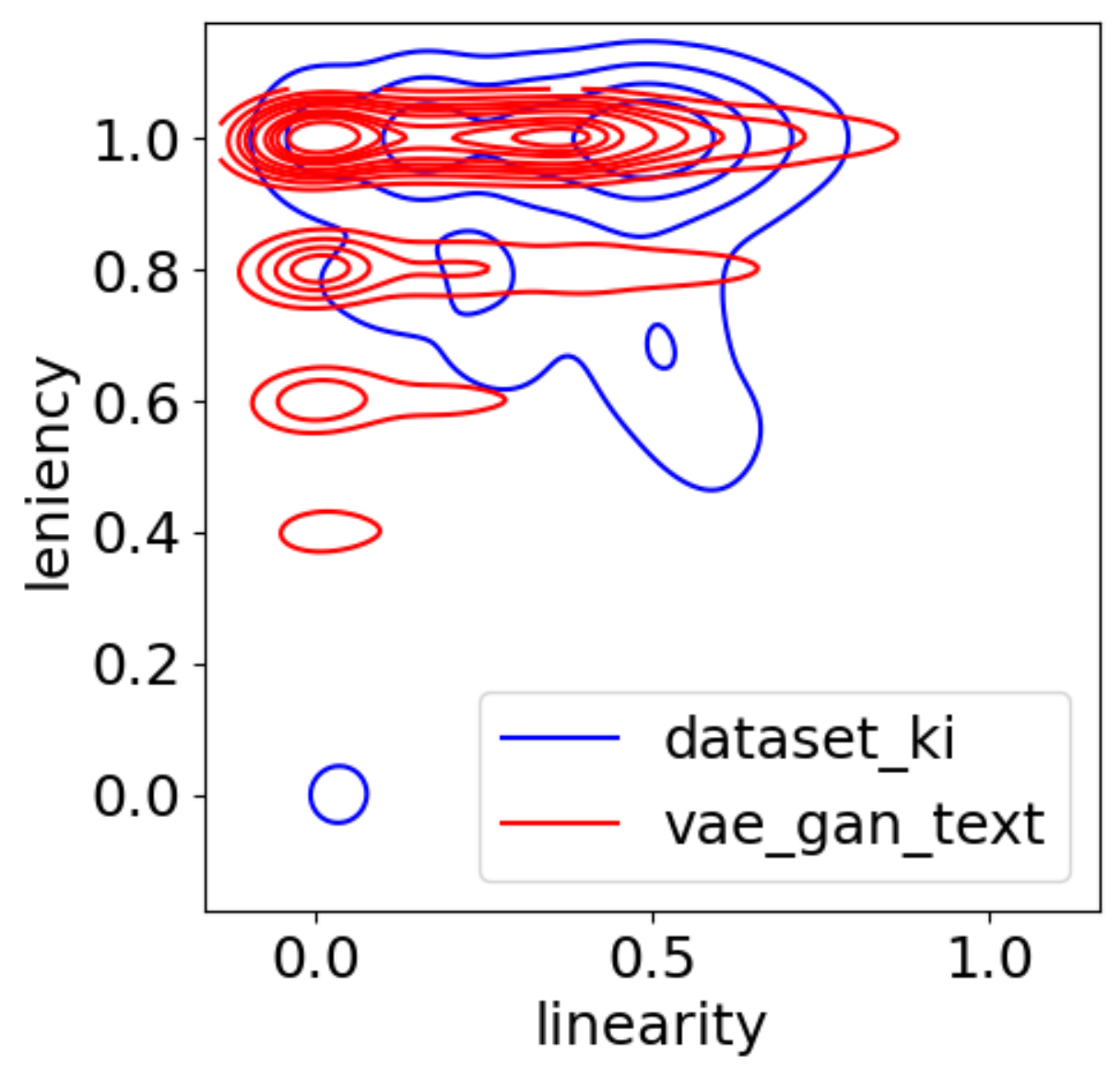}
         \caption{}
         \label{fig:ki_vae_gan_text}
     \end{subfigure}
     \centering
     \begin{subfigure}[b]{0.23\textwidth}
         \centering
         \includegraphics[width=\textwidth]{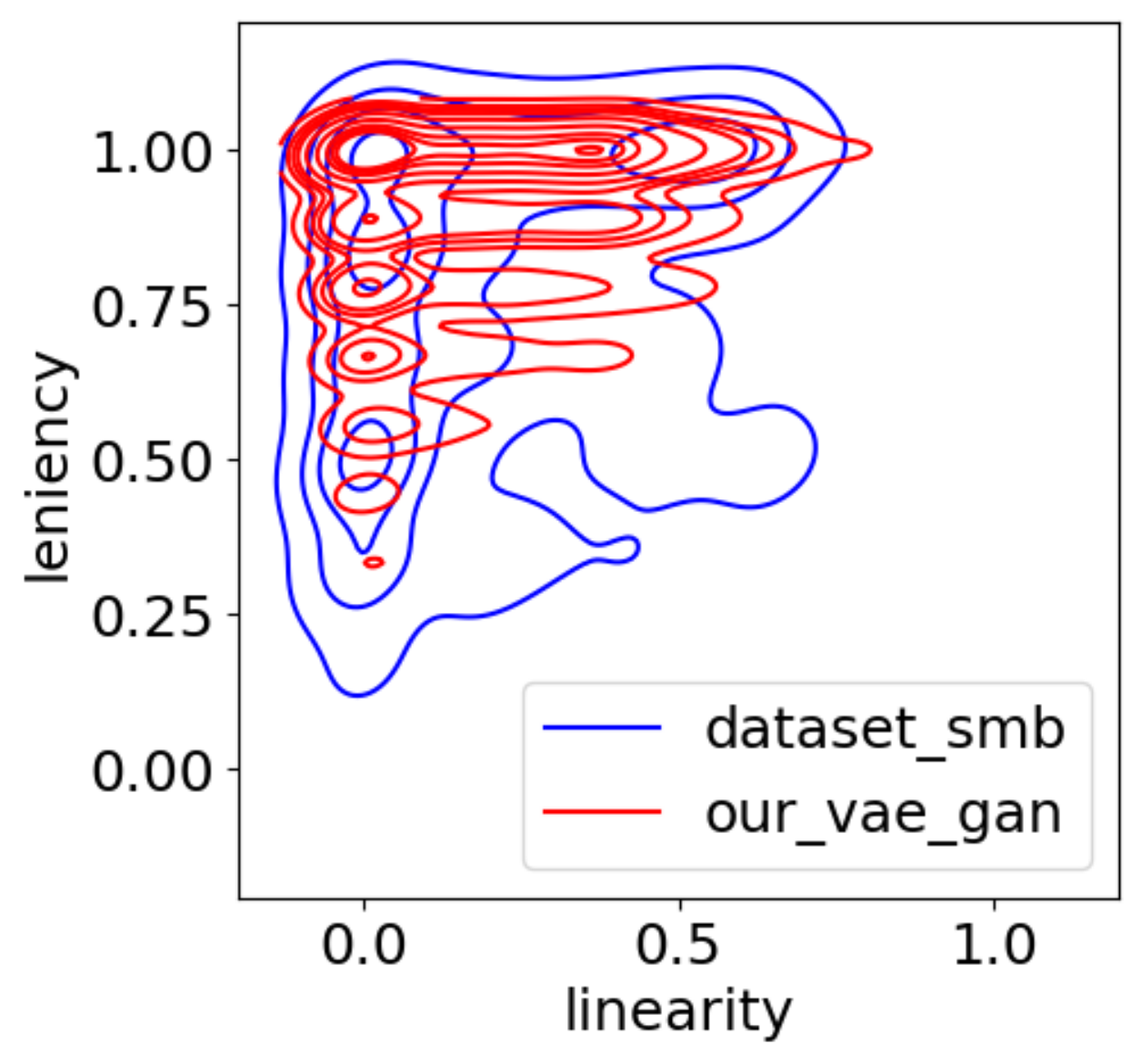}
         \caption{}
         \label{fig:smb_our_vae_gan}
     \end{subfigure}
     \begin{subfigure}[b]{0.23\textwidth}
         \centering
         \includegraphics[width=\textwidth]{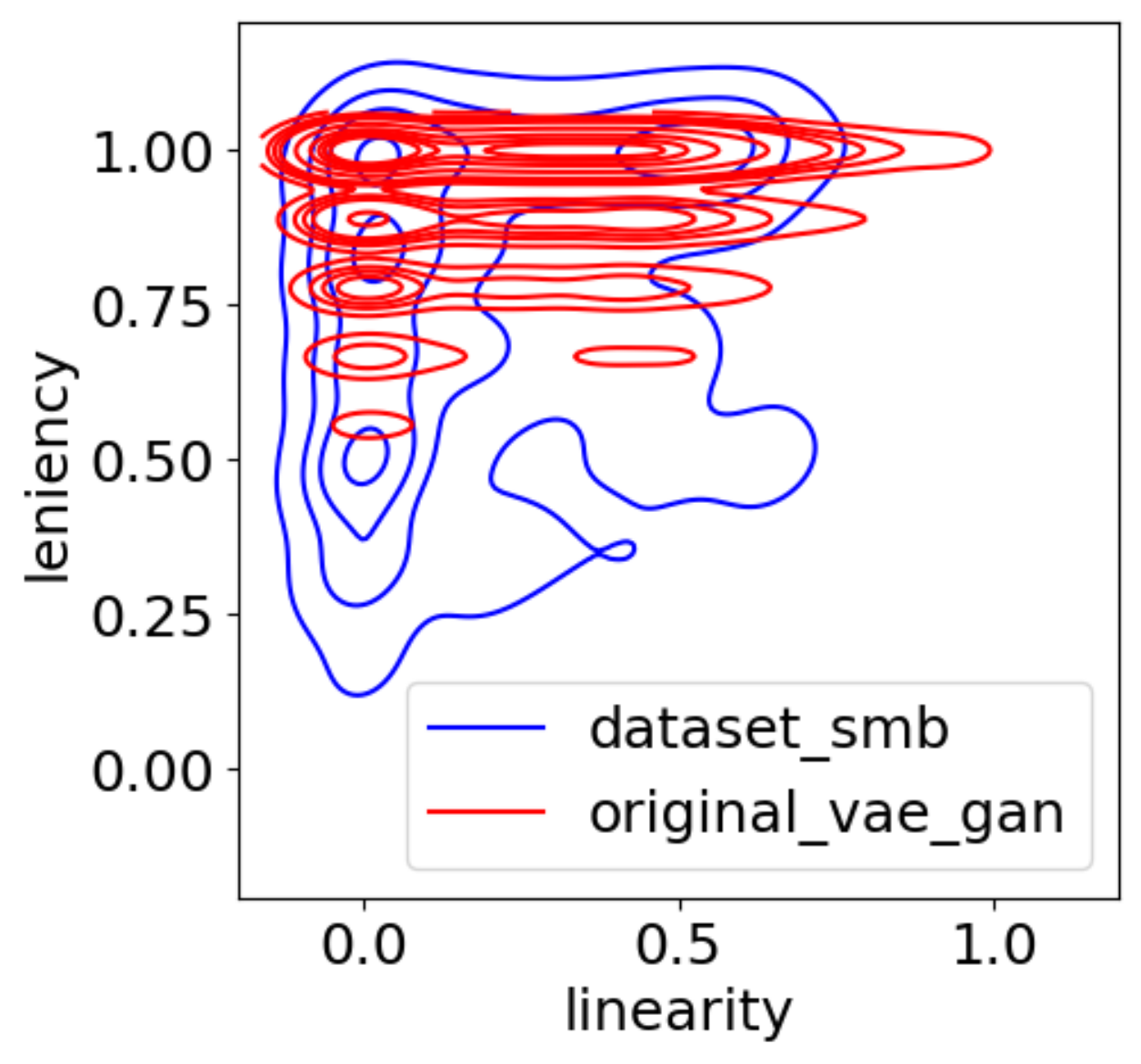}
         \caption{}
         \label{fig:smb_original_vae_gan}
     \end{subfigure}
     \begin{subfigure}[b]{0.23\textwidth}
         \centering
         \includegraphics[width=\textwidth]{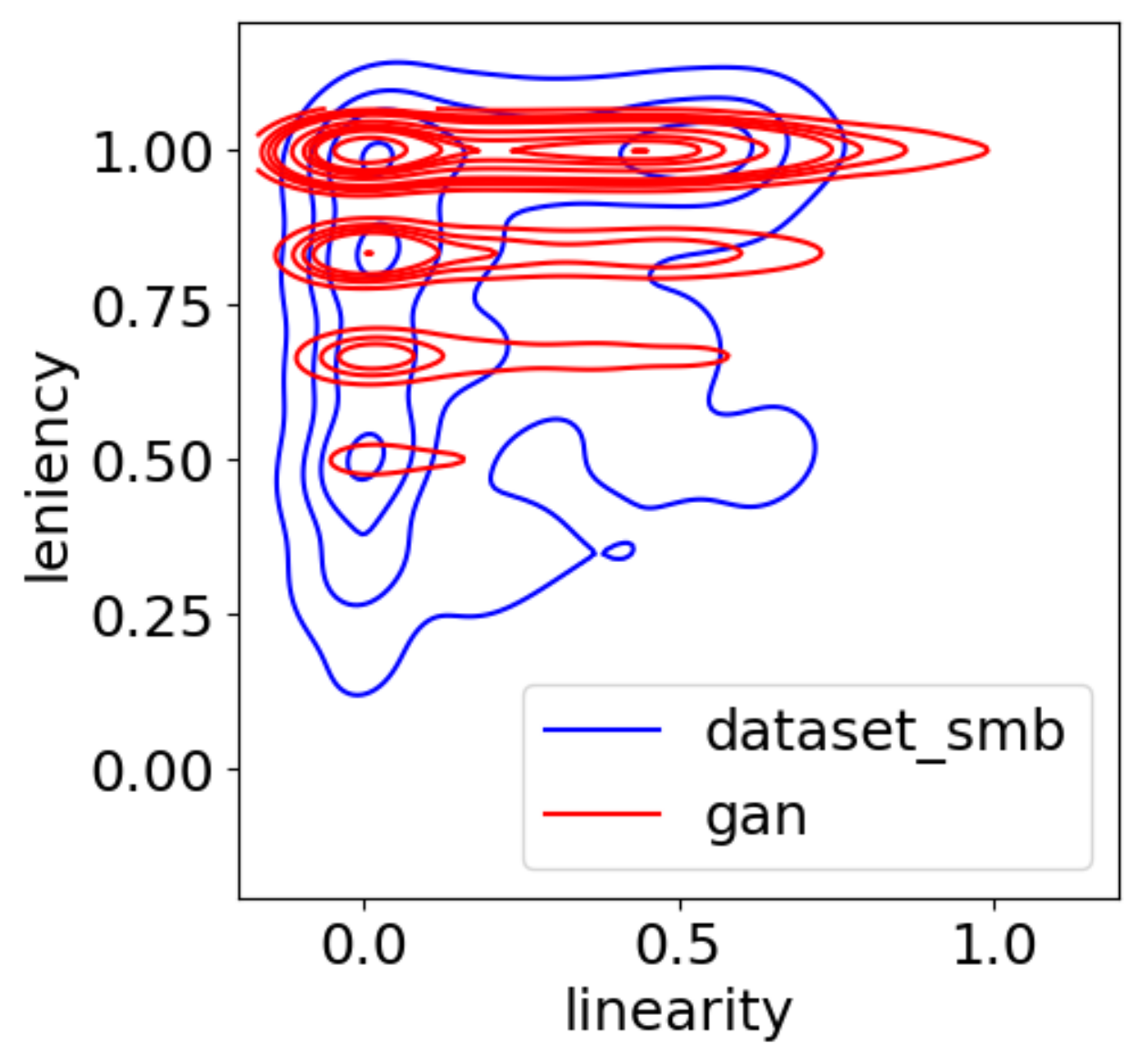}
         \caption{}
         \label{fig:smb_gan}
     \end{subfigure}
     \begin{subfigure}[b]{0.23\textwidth}
         \centering
         \includegraphics[width=\textwidth]{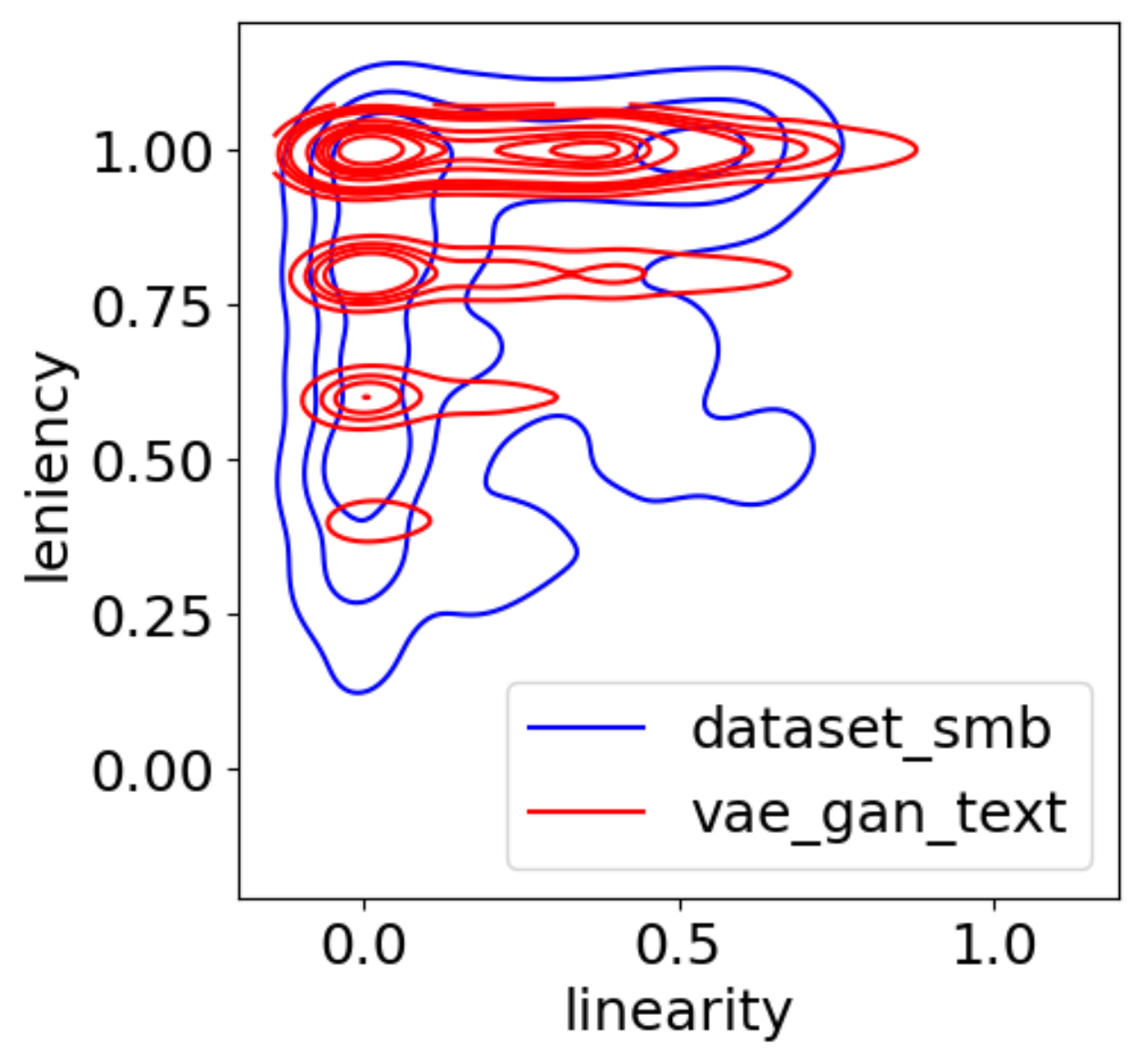}
         \caption{}
         \label{fig:smb_vae_gan_text}
     \end{subfigure}
        \caption{KDE plots of linearity vs. leniency for the top four models.}
        \label{fig:kde_plots}   
\end{figure*}

\begin{figure*}
     \centering
     \begin{subfigure}[b]{0.16\textwidth}
         \centering
         \includegraphics[width=\textwidth]{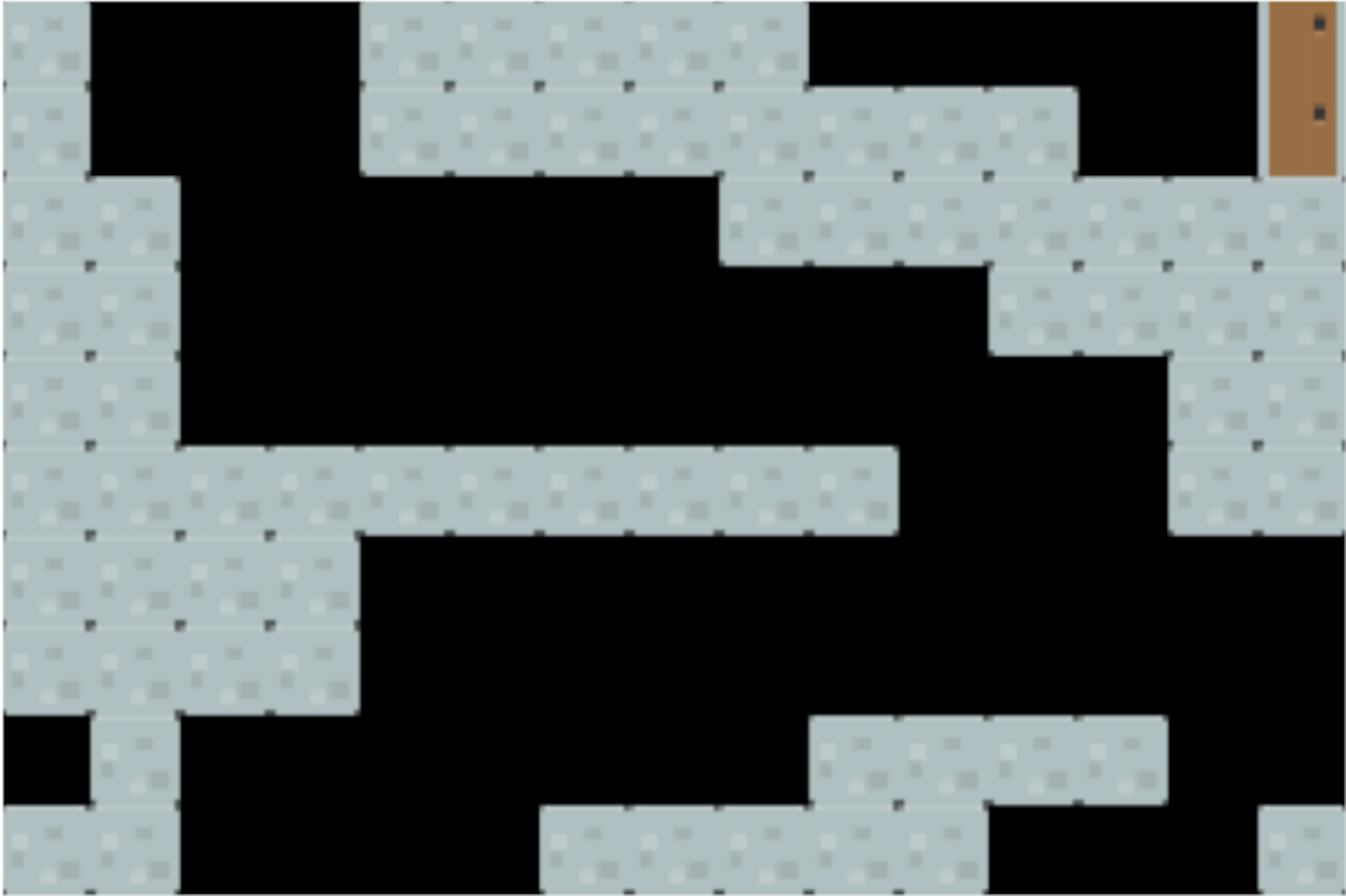}
         \caption{}
         \label{fig:gen_a}
     \end{subfigure}
     \begin{subfigure}[b]{0.16\textwidth}
         \centering
         \includegraphics[width=\textwidth]{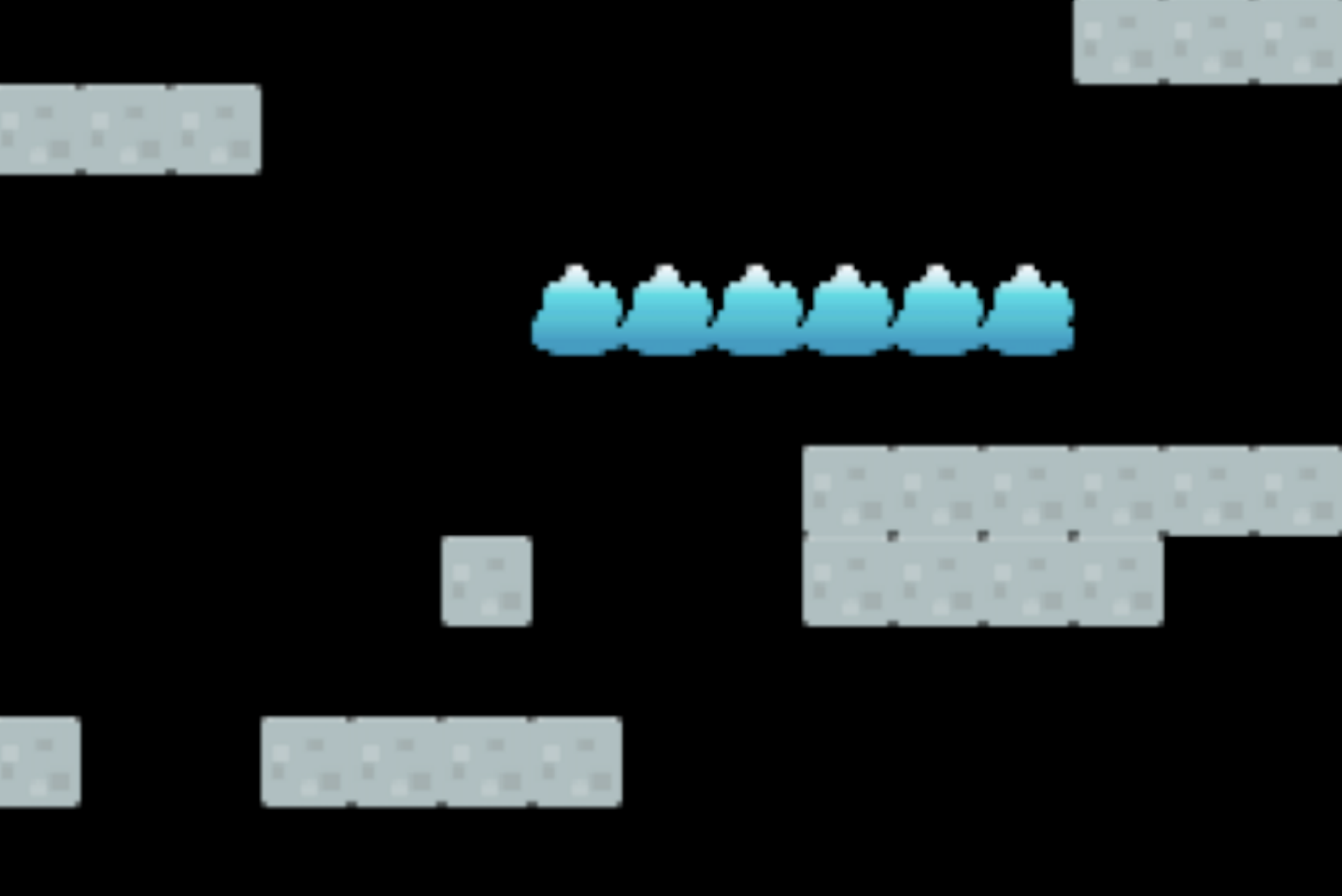}
         \caption{}
         \label{fig:gen_b}
     \end{subfigure}
     \begin{subfigure}[b]{0.16\textwidth}
         \centering
         \includegraphics[width=\textwidth]{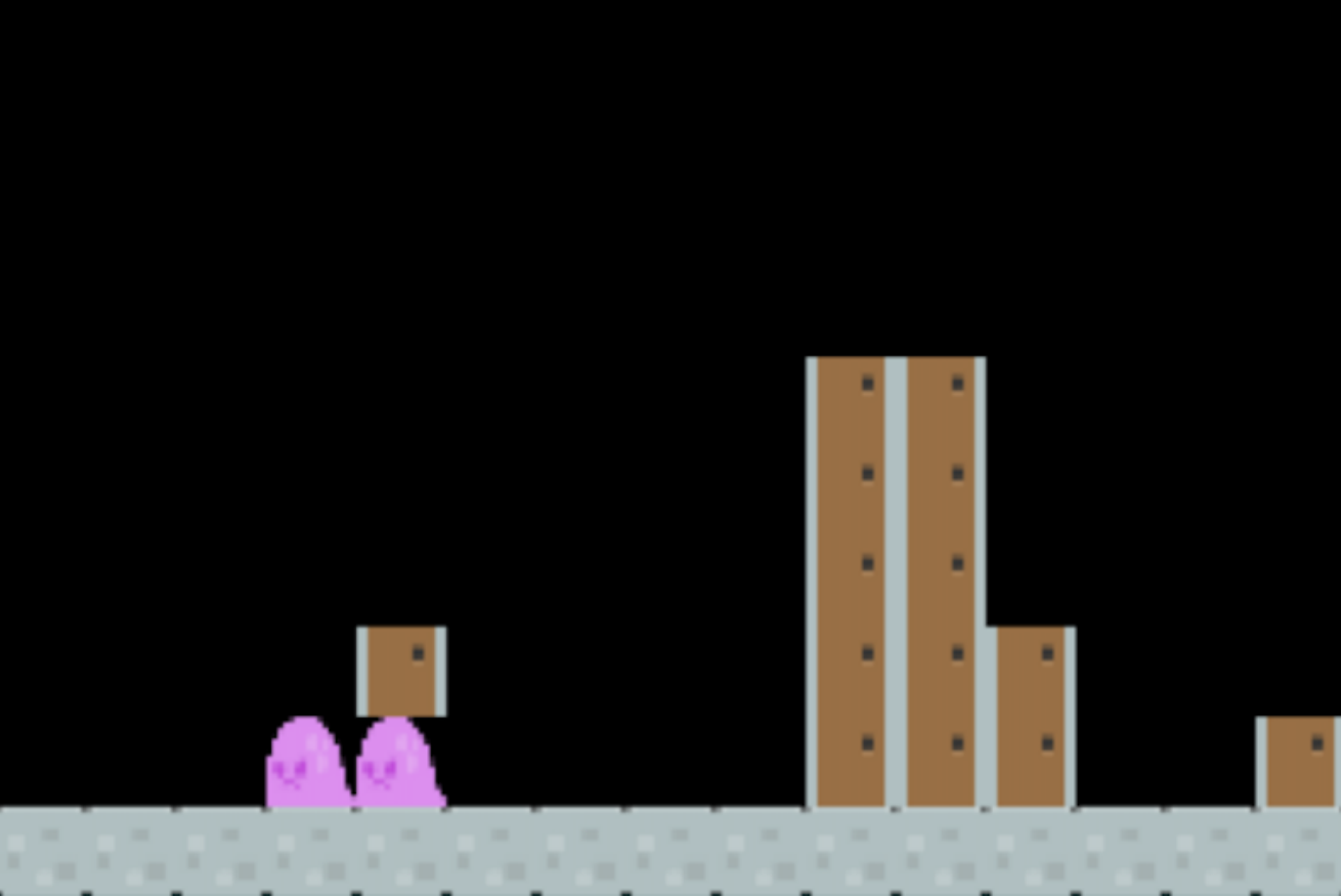}
         \caption{}
         \label{fig:gen_c}
     \end{subfigure}
     \begin{subfigure}[b]{0.16\textwidth}
         \centering
         \includegraphics[width=\textwidth]{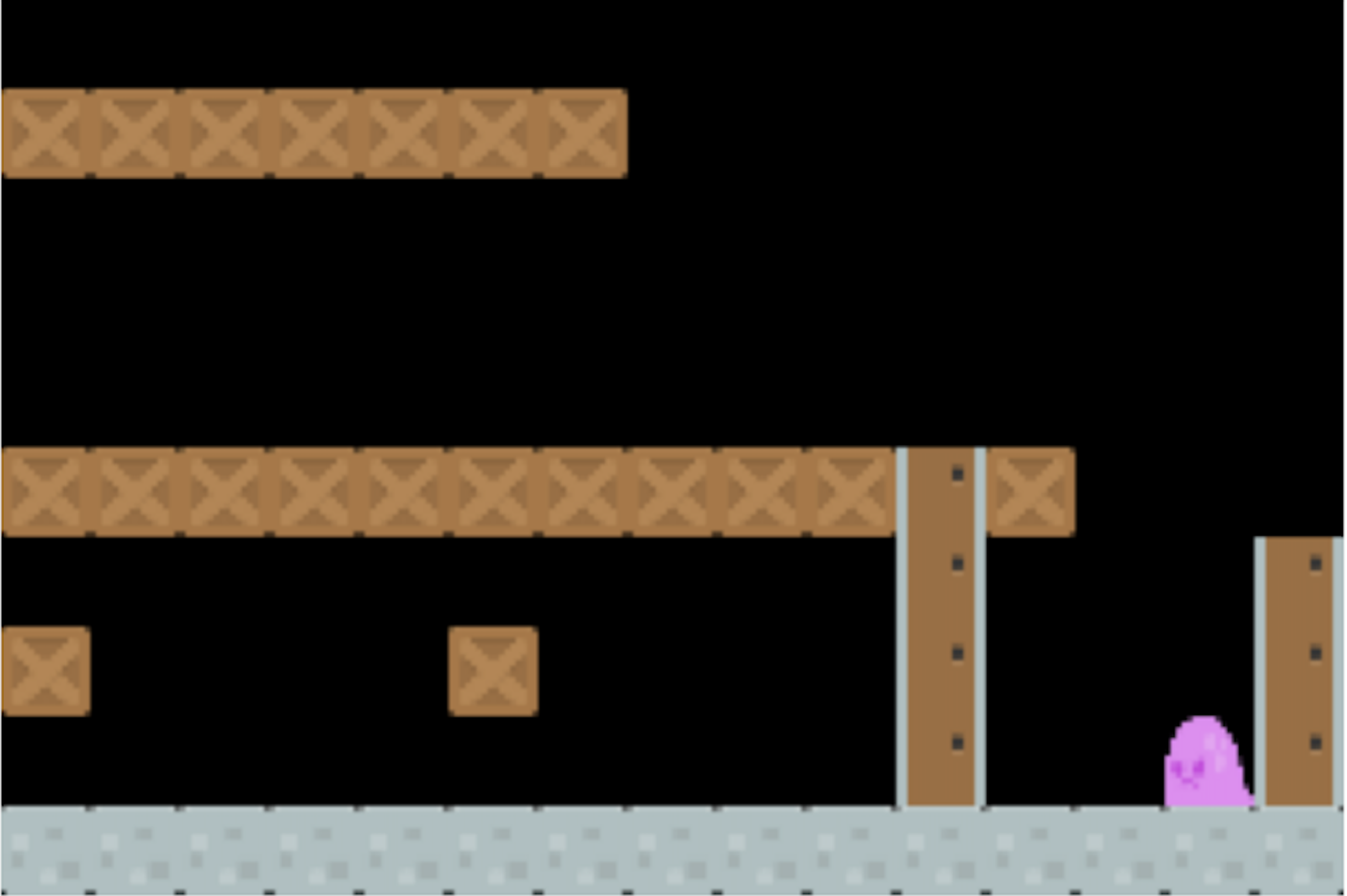}
         \caption{}
         \label{fig:gen_d}
     \end{subfigure}

     \centering
     \begin{subfigure}[b]{0.16\textwidth}
         \centering
         \includegraphics[width=\textwidth]{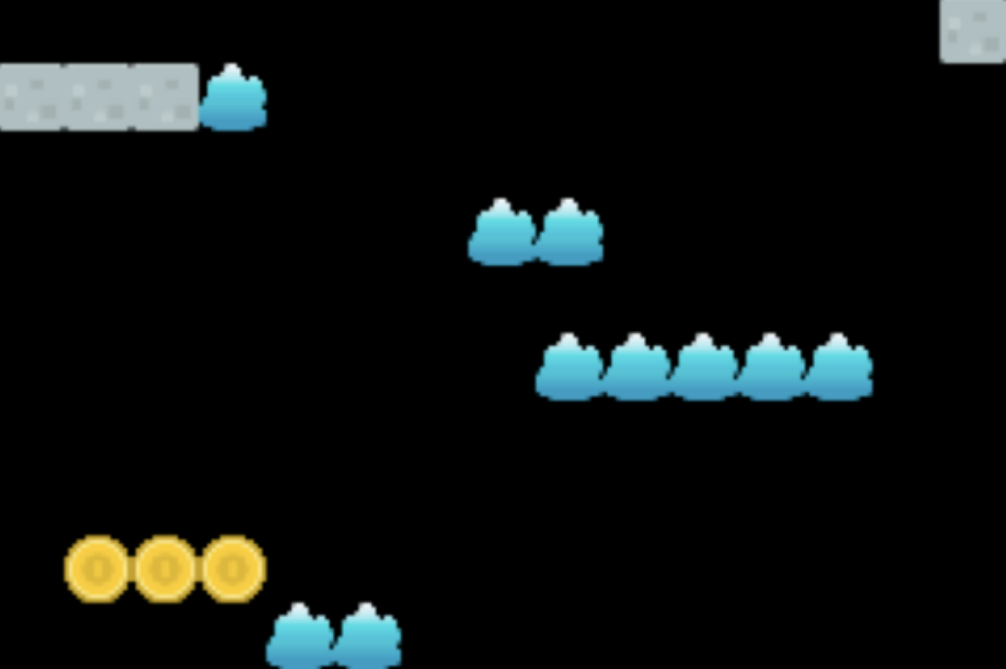}
         \caption{}
         \label{fig:b1}
     \end{subfigure}
     \begin{subfigure}[b]{0.16\textwidth}
         \centering
         \includegraphics[width=\textwidth]{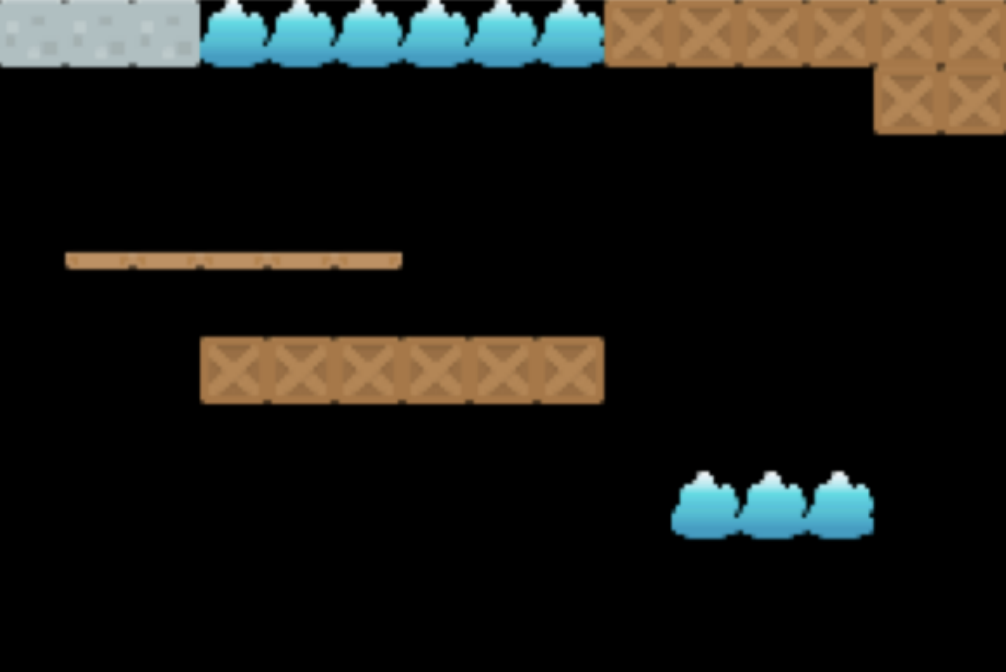}
         \caption{}
         \label{fig:b2}
     \end{subfigure}
     \begin{subfigure}[b]{0.16\textwidth}
         \centering
         \includegraphics[width=\textwidth]{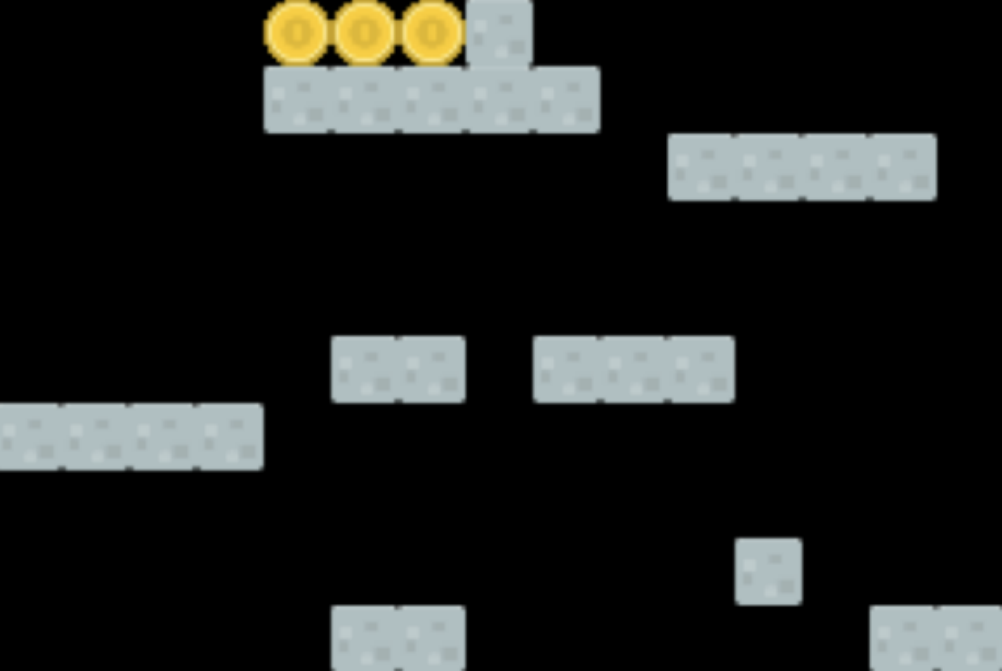}
         \caption{}
         \label{fig:b3}
     \end{subfigure}
     \begin{subfigure}[b]{0.16\textwidth}
         \centering
         \includegraphics[width=\textwidth]{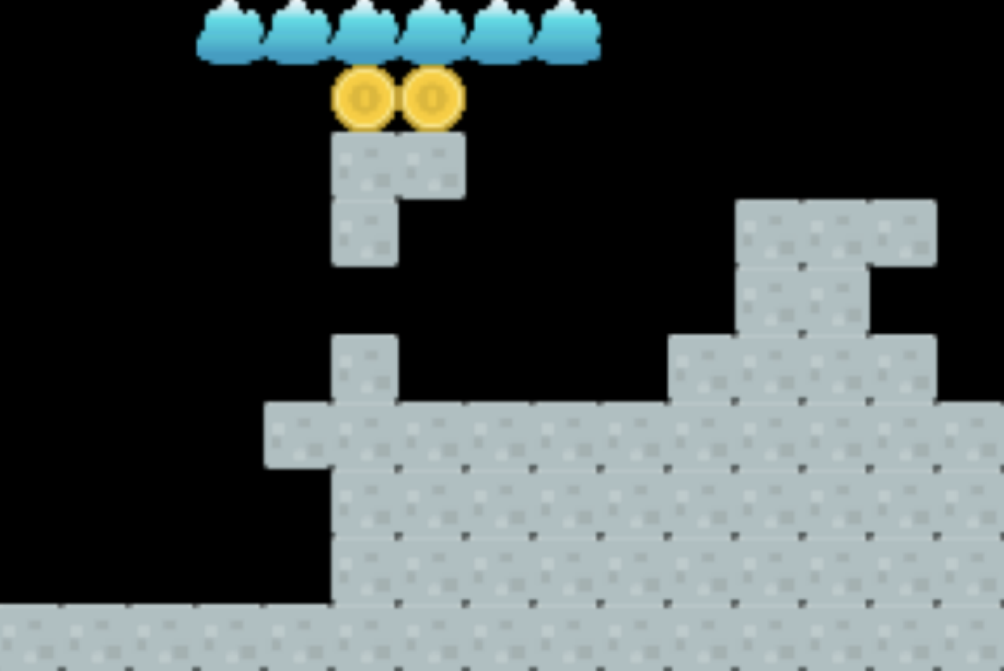}
         \caption{}
         \label{fig:b4}
     \end{subfigure}
        \caption{Sample generations of our VAE-GAN. Samples \ref{fig:gen_a} and \ref{fig:gen_b} look more similar to Kid Icarus levels, whereas samples \ref{fig:gen_c} and \ref{fig:gen_d} are closer to the Super Mario Bros. levels. Samples \ref{fig:b1}, \ref{fig:b2}, \ref{fig:b3}, \ref{fig:b4} in the second row blend the characteristics of both Kid Icarus and Super Mario Bros. levels.}
        \label{fig:sample_gens}
\end{figure*}

\section{Evaluation}\label {sec:eval}
%Similar one sentence purpose reminder as from the System Overview. 
%Ideal evaluation
%Say why that doesn't make sense (just an initial evaluation)
%What are we using as an approximation for evaluating ideal level %translation and level generation. 
%Cover in one sentence or less the different metrics (playability, %expressive range metrics, and accuracy). 
%Introduce the baselines and why we have them (again in like one %sentence)

The purpose of our framework is to accomplish simultaneous level generation and translation. The ideal way to evaluate the generative quality of a level generator is to conduct a human study, where designers/players inspect the generated outputs and assess them. However, this would be premature for an initial evaluation of this approach. Thankfully, researchers in the field of PCGML have proposed several metrics to address situations like this that do not require human evaluation \cite{smith2009rhythm, horn2014comparative, marino2015empirical, canossa2015towards, Summerville_2018}. On the other hand, we do not require human evaluation for the translation task as we can utilize metrics like accuracy and f1-score to assess the model's performance. 
The following subsections present the baselines and metrics we chose for assessing our framework.

\subsection{Baselines}
To evaluate the performance of our proposed framework, we used the following existing architectures for the level generation task:
\begin{itemize}
  \item \textbf{VAE-GAN}: We made use of the original VAE-GAN architecture \cite{larsen2016autoencoding}. Therefore, the primary difference between this baseline and our model is that this baseline only has one decoder/generator component instead of separate modules. Moreover, to make the comparison between architectures clearer, we leveraged the same reconstruction loss as our VAE-GAN. Noteworthy, we used $1e-6$ as the reconstruction vs. generation coefficient since this value was used in the original VAE-GAN paper \cite{larsen2016autoencoding}.
  \item \textbf{GAN}: For this baseline, we removed the VAE components from our framework and kept the generator and discriminator. The main difference between this GAN and our framework's GAN module is that it no longer has a shared layer.
  \item \textbf{VAE}: As above, but removing the GAN components of our framework.
  \item \textbf{VAE-GAN\_TEXT, VAE\_TEXT}: Same as the VAE-GAN and the VAE, but with the VGLC string representation of frames as input instead of frame images.
\end{itemize}

All of our baselines represent variations of our architecture, which will allow us to determine the utility of our VAE-GAN to this task in comparison to a typical VAE-GAN and the individual GAN and VAE components. In addition, we include variations of these baselines trained only on VGLC data to determine the utility of employing gameplay video data. We trained all our baselines using a batch size of $8$ for $300$ epochs, as empirical evidence showed that $300$ epochs was enough for our models to converge. We kept our baselines as similar as possible by utilizing the same hyperparameters as our model.

\subsection{Metrics}

Following previous research \cite{sarkar2021generating, Jadhav_Guzdial_2021}, we adapted the following metrics proposed by prior PCGML work to evaluate our framework's generation and translation quality:
\begin{itemize}
  \item \textbf{Linearity}: This metric measures how well a level conforms to a straight line. We employed the implementation of this metric from Jadhav and Guzdial \cite{Jadhav_Guzdial_2021}. 
  \item \textbf{Leniency}: This metric measures how hard a level is. We calculated this by subtracting the number of harmful/enemy tiles and half the number of moving tiles from the total number of tiles in the frame \cite{sarkar2021generating}.
  \item \textbf{Interestingness}: This metric measures the number of interesting tiles in a level. To calculate this, we counted the number of doors, moving platforms, collectibles, and harmful tiles \cite{sarkar2021generating}.
  \item \textbf{Playability}: The purpose of this metric is to determine whether a level segment is playable or not by trying to find a path from the lowest piece of level structure to the highest piece of level structure. We acknowledge this is imperfect and is only meant as a rough estimation. We adapted this implementation of playability and pathfinding agents from Sarkar and Cooper \cite{sarkar2021dungeon, sarkar2021generating}.
  \item \textbf{Accuracy}: We used this metric to evaluate our model's performance regarding the ability to translate level segments. We calculated accuracy by dividing the number of correctly translated tiles by the total number of tiles in the level segment. 
\end{itemize}
Additionally, we used the following metrics as means of comparing our model and baselines based on the values that we obtained using the metrics mentioned above:
\begin{itemize}
    \item \textbf{Energy Distance (E-Distance)}: The purpose of this function is to measure the distance between two distributions \cite{rizzo2016energy}. In this paper, we considered the linearity, leniency, and interestingness of generated samples as features of their distribution to estimate how similar the outputs of each model are to our dataset. We used the GeomLoss library \cite{feydy2019interpolating} to calculate this metric.
    \item \textbf{Kernel Density Estimation (KDE)}: This statistical approach is used to estimate the probability density function of a distribution based on a set of samples \cite{rosenblatt1956remarks}. We utilized KDE plots to visualize the generative space of our models and compare them against both our training and test sets. 
\end{itemize}

We note that to calculate the training metrics, we generated $3956$ random samples for each model to have an equal number of samples as our training dataset. Similarly, we generated $360$ random samples to compare our model's outputs against the test dataset.

\begin{figure*}
     \centering
     \begin{subfigure}[b]{0.22\textwidth}
         \centering
         \includegraphics[width=\textwidth]{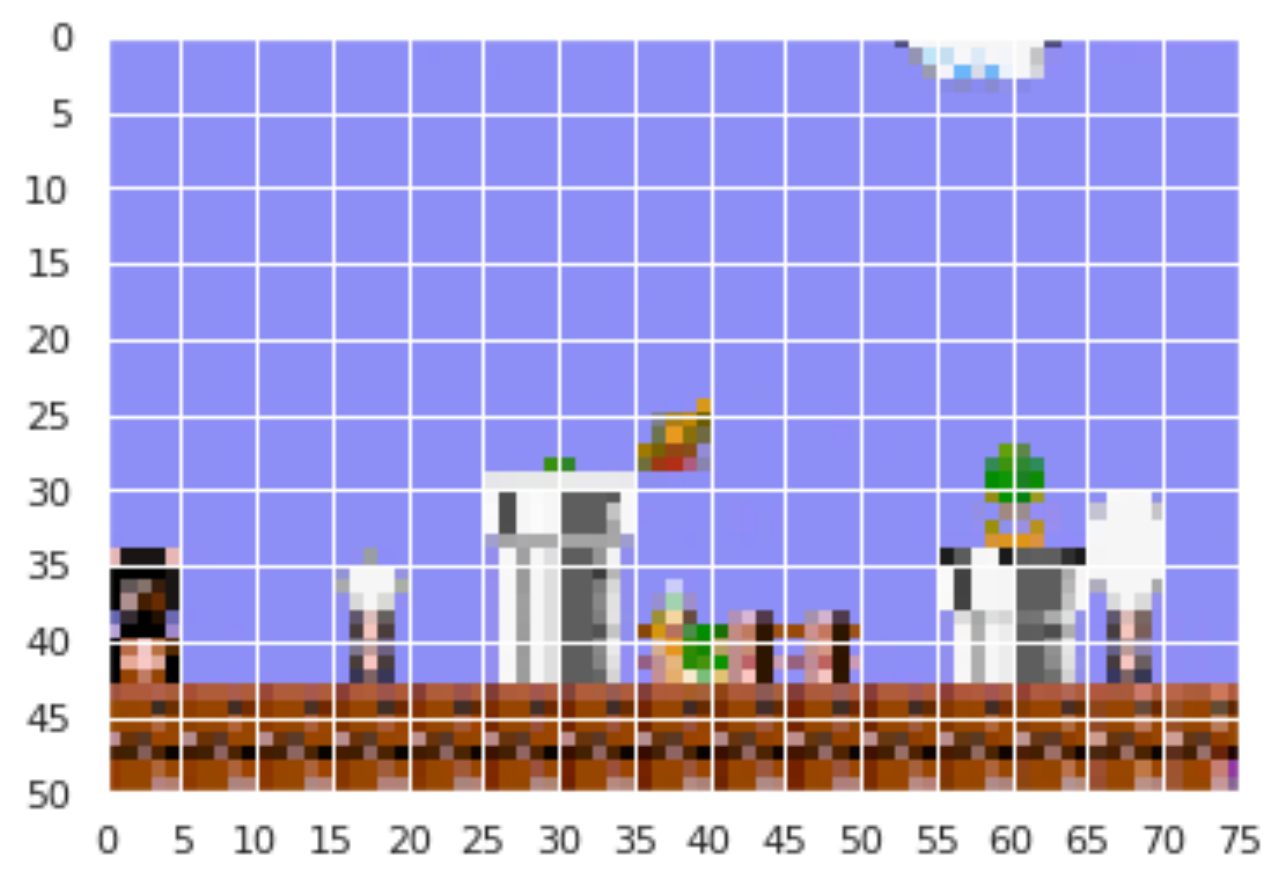}
         \caption{}
         \label{fig:to_trans_1}
     \end{subfigure}
     \begin{subfigure}[b]{0.22\textwidth}
         \centering
         \includegraphics[width=\textwidth]{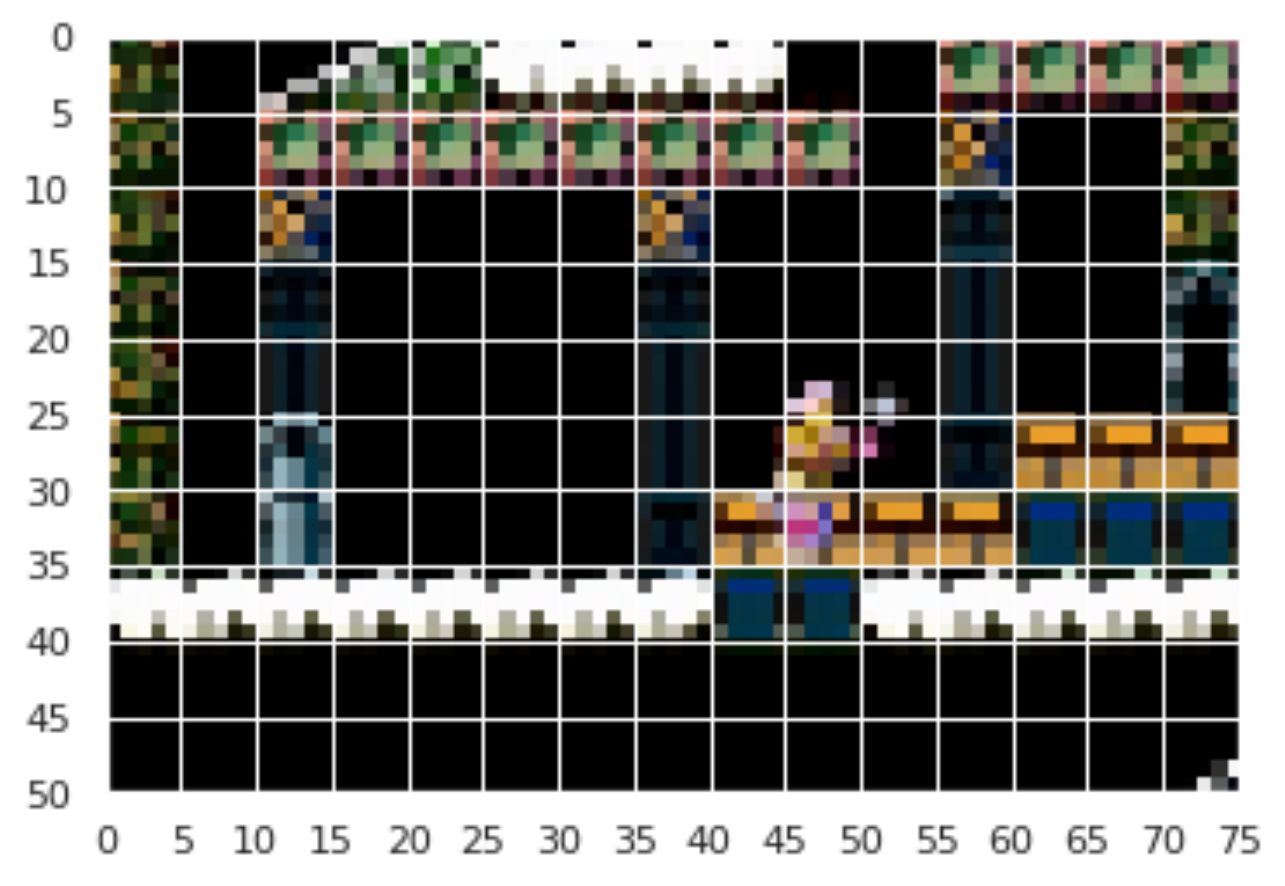}
         \caption{}
         \label{fig:to_trans_2}
     \end{subfigure}
     \begin{subfigure}[b]{0.22\textwidth}
         \centering
         \includegraphics[width=\textwidth]{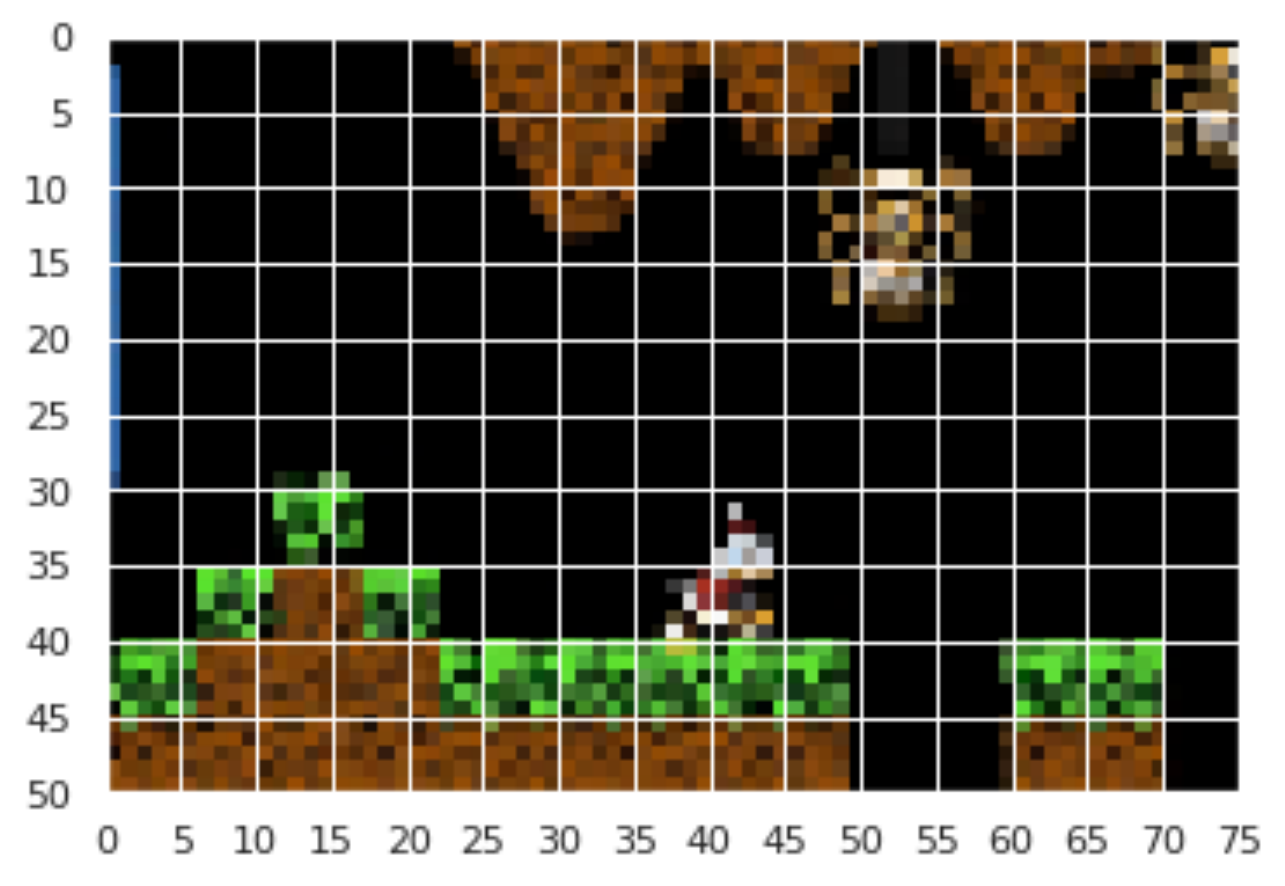}
         \caption{}
         \label{fig:to_trans_3}
     \end{subfigure}
     \begin{subfigure}[b]{0.22\textwidth}
         \centering
         \includegraphics[width=\textwidth]{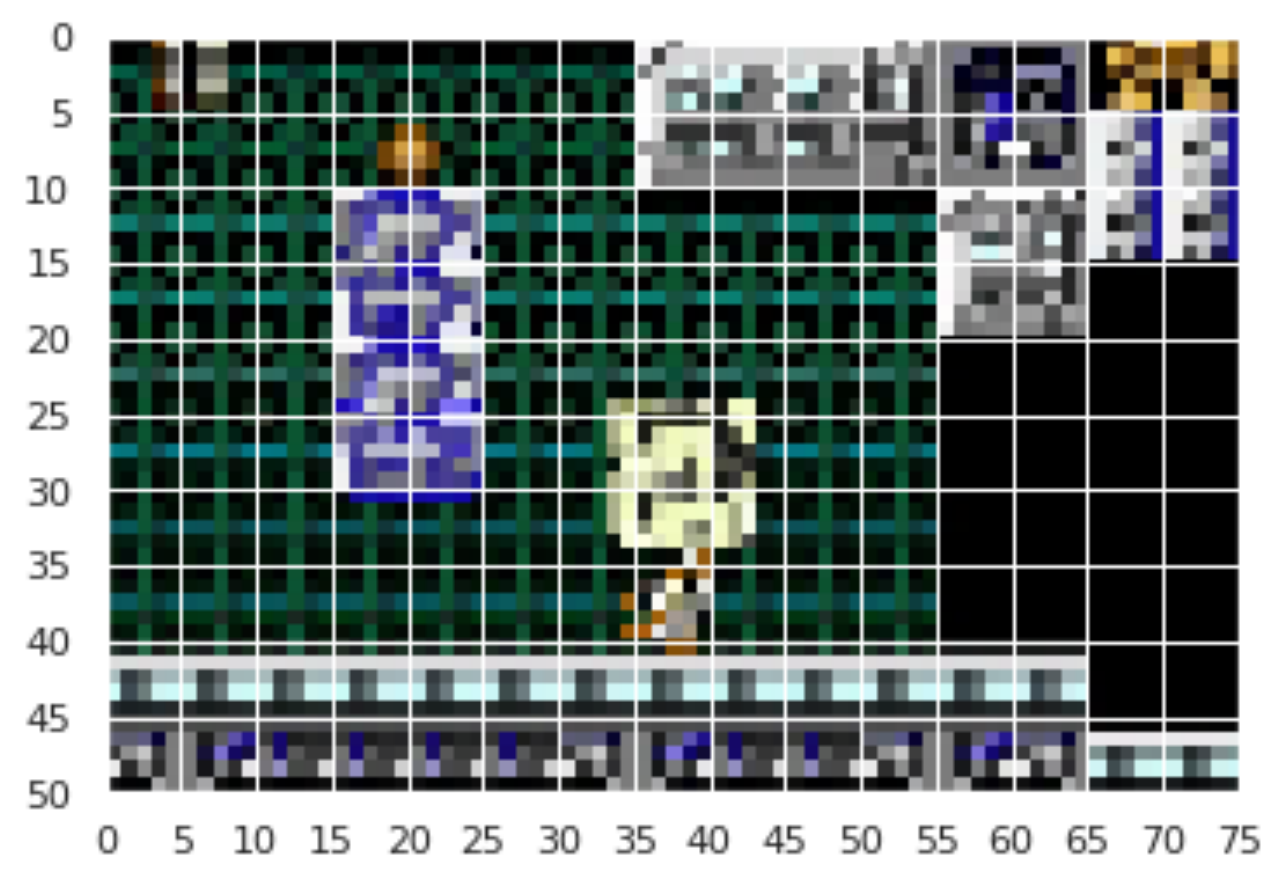}
         \caption{}
         \label{fig:to_trans_4}
     \end{subfigure}
     \centering
     \begin{subfigure}[b]{0.22\textwidth}
         \centering
         \includegraphics[scale=0.70, keepaspectratio]{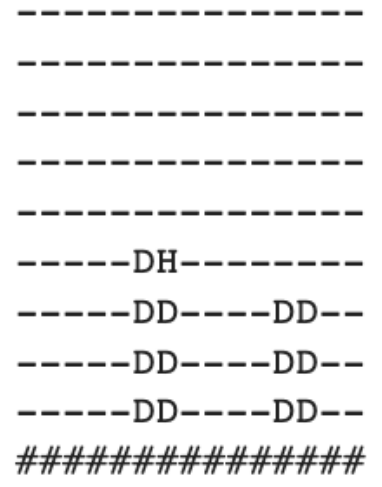}
         \caption{}
         \label{fig:trans_1}
     \end{subfigure}
     \begin{subfigure}[b]{0.22\textwidth}
         \centering
         \includegraphics[scale=0.70, keepaspectratio]{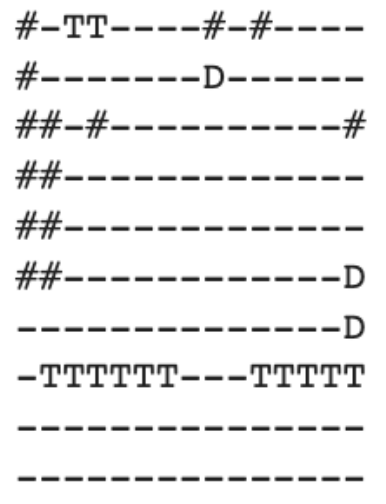}
         \caption{}
         \label{fig:trans_2}
     \end{subfigure}
     \begin{subfigure}[b]{0.22\textwidth}
         \centering
         \includegraphics[scale=0.70, keepaspectratio]{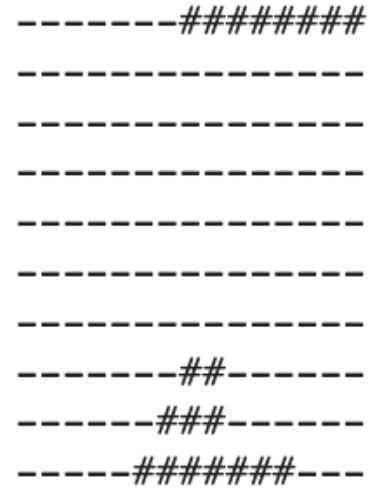}
         \caption{}
         \label{fig:trans_3}
     \end{subfigure}
     \begin{subfigure}[b]{0.22\textwidth}
         \centering
         \includegraphics[scale=0.70, keepaspectratio]{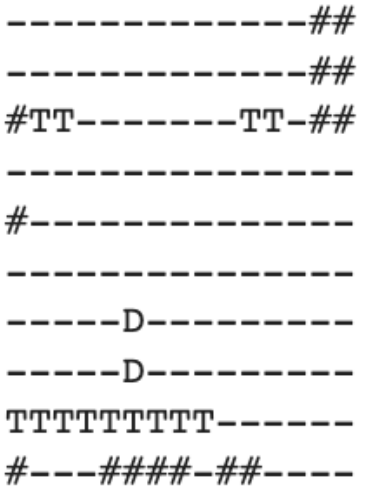}
         \caption{}
         \label{fig:trans_4}
     \end{subfigure}
        \caption{Sample translations of our VAE-GAN. Figures \ref{fig:to_trans_1} (Super Mario Bros.) and \ref{fig:to_trans_2} (Kid Icarus) belong to the test set of our dataset, and figures \ref{fig:to_trans_3} (Duck Tales), and \ref{fig:to_trans_4} (Megaman) are sample frames from unseen games.}
        \label{fig:sample_translations}
\end{figure*}

\section{Results} \label{sec:results} 
% TABLE walkthrough %
Table \ref{table:table_1} presents the results of the evaluation metrics for our framework and all baselines. We bold the best value for a particular column or metric, and a dash indicates that the corresponding model does not have a defined value for that metric. %Moreover, the use of \_TEXT in model names indicates that the model was trained with the VGLC string representation of frames as input instead of frame images.%
Notably, our table has two playability columns, as we ran Super Mario Bros. and Kid Icarus pathfinding agents for all generated outputs. We made this decision because our model and baselines learn a single latent space, so they produce outputs covering both games and the space between them.
We observe that our VAE-GAN outperforms all the baselines regarding e-distance across both training and test datasets. This suggests that the generated outputs from our VAE-GAN are closer to the original datasets than the other baselines. Also, the test e-distance results provide evidence for our model's superior ability to generalize. The test e-distance also demonstrates evidence of overfitting, as the text-based models perform poorly on it. We take this as evidence that using gameplay video is beneficial for the level generation task. A possible explanation for this is that drawing on video data could have a similar effect as data augmentation in terms of introducing small variations on the same structures. For example, each second of Super Mario Bros. gameplay video produces two frames in our dataset (with $FPS=2$). This results in some frame images with significant overlap, but with differences in their pixel representation due to the dynamic nature of gameplay video (e.g., movement of the player and enemies). In comparison, the static string representation had no such variation and no way to generate it automatically.
In terms of reconstruction accuracy, our framework achieves the best performance, tied with the VAE\_TEXT framework, though it is clearly superior in terms of generation quality. 

Turning our attention to the playability results, we see that all the playability (SMB) percentages are quite close except for the VAE baseline. By inspecting this baseline's outputs, we recognized that its outputs had less variety and simpler structures, leading to improved playability but an inferior e-distance. Furthermore, the playability (KI) results are similar and rather low. We anticipate that this is due to this Kid Icarus pathfinding agent \cite{sarkar2021dungeon, sarkar2021generating} only achieving $77\%$ on the real Kid Icarus level segments of our dataset. The original VAE-GAN does outperform our approach for playability (KI) but not in terms of playability (SMB) or other metrics. We take this as evidence of the utility of the VAE-GAN architecture generally and of our alterations to this architecture specifically.

%KDE Plot Walkthrough%
Figure \ref{fig:kde_plots} demonstrates linearity vs. leniency KDE plots for the top four models in terms of training e-distance. The first row of Figure \ref{fig:kde_plots} plots the linearity vs. leniency against all the Kid Icarus samples from our training dataset for our VAE-GAN, the original VAE-GAN, GAN, and the VAE-GAN\_TEXT models, respectively. The second row plots the same metrics for the same models against our dataset's Super Mario Bros. samples. We chose these architectures as they had the best results in Table \ref{table:table_1}. As we move from left to right in each row, we see a decreasing trend regarding coverage of the original distribution. We also observe that this decreasing trend of coverage correlates with the trend that we see in the e-distance values of these four models. We refer the readers to the supplementary material of this paper for an overview of our KDE plots on the test dataset.

%Generations Walkthrough%
Figure \ref{fig:sample_gens} displays four instances of our VAE-GAN's generated outputs using a common tile representation (obtained from Kenney \footnote{kenney.nl/assets/platformer-art-deluxe}). Samples \ref{fig:gen_a} and \ref{fig:gen_b} have a similar structure as the Kid Icarus levels, while samples \ref{fig:gen_c} and \ref{fig:gen_d} have a closer resemblance to the Super Mario Bros. levels. Moreover, the sample images in the second row blend the characteristics of both games. We should note that we looked at a hundred randomly generated outputs to find the examples that were closest to these criteria. To elaborate on the blended aspects of examples in the second row, although sample \ref{fig:b1} has solid-top platforms (cloud-like tiles) that are placed in a manner that is similar to Kid Icarus level structure, it has collectibles (coin-shaped tiles) that only occur in Super Mario Bros. levels. This observation indicates that our model can blend the aspects of both games to generate novel level segments that were not included in the training dataset.

% Translation Walkthrough$
Figure \ref{fig:sample_translations} presents sample translations of our framework. The upper row consists of frames from Super Mario Bros., Kid Icarus, Duck Tales, and Megaman, respectively. The translation of each frame is below it. We observe that the translation performance is significantly better for the first two samples (\ref{fig:to_trans_1}, \ref{fig:to_trans_2}) since our model saw samples from these two games during training. We selected outputs to demonstrate this, but note from Table \ref{table:table_1} that we had high translation accuracy.

\section{Limitations and Future Work} \label{sec:future}
% can identify limitations of the current research and propose research that could address them. In the “Limitations” section, authors may include a discussion of where their current approach falls short in terms of fully solving or addressing the problem(s) or lack(s) from the Introduction.
In this paper, we aimed to develop a framework that is able to perform level generation and translation. Our initial evaluations suggest that our VAE-GAN architecture was able to learn both of these tasks and achieve better results compared to other frequently-used architectures in PCGML. However, there are multiple avenues for improvement. To begin with, the GAN component of our model is not controllable, meaning that we cannot select which type of game level it generates. Therefore, one approach for improvement is to develop a conditional version of this framework \cite{mirza2014conditional}. 
Second, we trained and evaluated our model with data from two games. However, expanding the training dataset to include more games may improve our framework's ability to better generalize over unseen game frames. Furthermore, this opens up the possibility of generating new level segments for unseen games. To be clear, if the framework is able to translate unseen level segments effectively, one can take any gameplay video of a new 2D platformer game and feed its frames through the VAE to attain their translations. Then, these (frame, translation) pairs can be used to generate novel level segments for that new game. 

 Our framework's effectiveness may have been impacted by the limitations of the annotated dataset. For instance, ‌‌‌‌each moving platform in Kid Icarus has a specific length. However, all the moving platforms in the VGLC representation take up a whole row. As a result, there is a mismatch between frame images with moving platforms and their corresponding VGLC translation. Moreover, the VGLC simplifies the representation of game levels in the sense that different game objects may be defined with the same symbol. For example, Goombas, Piranha Plants, and Koopas in Super Mario Bros. are all represented with the same character in the VGLC representation, despite the fact that they behave differently in game. Employing a different level representation could help avoid these issues.

\section{Conclusions} \label{sec:conc}
% a summary of your work now under the assumption that the reader has read your whole paper. I often recommend students include the 2-4 big points you most want readers to remember from the paper (even if they’ve forgotten everything else).
In this paper, we proposed a novel framework for simultaneous level segment generation and translation. To achieve this goal, we trained a novel VAE-GAN-based architecture on a dataset of two human-annotated platformer games. By comparing our framework against multiple baselines, we showed evidence that learning these two tasks jointly can lead to an overall better performance in terms of generation quality and translation accuracy.

\vspace{12pt}

\bibliography{conference_101719}

\begin{thebibliography}{10}
\providecommand{\url}[1]{#1}
\csname url@samestyle\endcsname
\providecommand{\newblock}{\relax}
\providecommand{\bibinfo}[2]{#2}
\providecommand{\BIBentrySTDinterwordspacing}{\spaceskip=0pt\relax}
\providecommand{\BIBentryALTinterwordstretchfactor}{4}
\providecommand{\BIBentryALTinterwordspacing}{\spaceskip=\fontdimen2\font plus
\BIBentryALTinterwordstretchfactor\fontdimen3\font minus
  \fontdimen4\font\relax}
\providecommand{\BIBforeignlanguage}[2]{{%
\expandafter\ifx\csname l@#1\endcsname\relax
\typeout{** WARNING: IEEEtran.bst: No hyphenation pattern has been}%
\typeout{** loaded for the language `#1'. Using the pattern for}%
\typeout{** the default language instead.}%
\else
\language=\csname l@#1\endcsname
\fi
#2}}
\providecommand{\BIBdecl}{\relax}
\BIBdecl

\bibitem{summerville2018procedural}
A.~Summerville, S.~Snodgrass, M.~Guzdial, C.~Holmg{\aa}rd, A.~K. Hoover,
  A.~Isaksen, A.~Nealen, and J.~Togelius, ``Procedural content generation via
  machine learning (pcgml),'' \emph{IEEE Transactions on Games}, vol.~10,
  no.~3, pp. 257--270, 2018.

\bibitem{Jadhav_Guzdial_2021}
\BIBentryALTinterwordspacing
M.~Jadhav and M.~Guzdial, ``Tile embedding: A general representation for level
  generation,'' \emph{Proceedings of the AAAI Conference on Artificial
  Intelligence and Interactive Digital Entertainment}, vol.~17, no.~1, pp.
  34--41, Oct. 2021. [Online]. Available:
  \url{https://ojs.aaai.org/index.php/AIIDE/article/view/18888}
\BIBentrySTDinterwordspacing

\bibitem{opencv_library}
G.~Bradski, ``{The OpenCV Library},'' \emph{Dr. Dobb's Journal of Software
  Tools}, 2000.

\bibitem{Guzdial_Riedl_2021}
\BIBentryALTinterwordspacing
M.~Guzdial and M.~Riedl, ``Game level generation from gameplay videos,''
  \emph{Proceedings of the AAAI Conference on Artificial Intelligence and
  Interactive Digital Entertainment}, vol.~12, no.~1, pp. 44--50, Jun. 2021.
  [Online]. Available:
  \url{https://ojs.aaai.org/index.php/AIIDE/article/view/12861}
\BIBentrySTDinterwordspacing

\bibitem{summerville2016vglc}
A.~J. Summerville, S.~Snodgrass, M.~Mateas, and S.~Ontan{\'o}n, ``The vglc: The
  video game level corpus,'' \emph{arXiv preprint arXiv:1606.07487}, 2016.

\bibitem{deng2009imagenet}
J.~Deng, W.~Dong, R.~Socher, L.-J. Li, K.~Li, and L.~Fei-Fei, ``Imagenet: A
  large-scale hierarchical image database,'' in \emph{2009 IEEE conference on
  computer vision and pattern recognition}.\hskip 1em plus 0.5em minus
  0.4em\relax Ieee, 2009, pp. 248--255.

\bibitem{liu2015deep}
Z.~Liu, P.~Luo, X.~Wang, and X.~Tang, ``Deep learning face attributes in the
  wild,'' in \emph{Proceedings of the IEEE international conference on computer
  vision}, 2015, pp. 3730--3738.

\bibitem{NEURIPS2021_2bcab9d9}
\BIBentryALTinterwordspacing
D.~Smirnov, M.~GHARBI, M.~Fisher, V.~Guizilini, A.~Efros, and J.~M. Solomon,
  ``Marionette: Self-supervised sprite learning,'' in \emph{Advances in Neural
  Information Processing Systems}, M.~Ranzato, A.~Beygelzimer, Y.~Dauphin,
  P.~Liang, and J.~W. Vaughan, Eds., vol.~34.\hskip 1em plus 0.5em minus
  0.4em\relax Curran Associates, Inc., 2021, pp. 5494--5505. [Online].
  Available:
  \url{https://proceedings.neurips.cc/paper/2021/file/2bcab9d935d219641434683dd9d18a03-Paper.pdf}
\BIBentrySTDinterwordspacing

\bibitem{Chen_Sydora_Burega_Mahajan_Abdullah_Gallivan_Guzdial_2020}
\BIBentryALTinterwordspacing
E.~Chen, C.~Sydora, B.~Burega, A.~Mahajan, A.~Abdullah, M.~Gallivan, and
  M.~Guzdial, ``Image-to-level: Generation and repair,'' \emph{Proceedings of
  the AAAI Conference on Artificial Intelligence and Interactive Digital
  Entertainment}, vol.~16, no.~1, pp. 189--195, Oct. 2020. [Online]. Available:
  \url{https://ojs.aaai.org/index.php/AIIDE/article/view/7429}
\BIBentrySTDinterwordspacing

\bibitem{snodgrass2016approach}
S.~Snodgrass and S.~Ontanon, ``An approach to domain transfer in procedural
  content generation of two-dimensional videogame levels,'' in
  \emph{Proceedings of the AAAI Conference on Artificial Intelligence and
  Interactive Digital Entertainment}, vol.~12, no.~1, 2016, pp. 79--85.

\bibitem{sarkar19controllable}
A.~Sarkar, Z.~Yang, and S.~Cooper, ``Controllable level blending between games
  using variational autoencoders,'' in \emph{Proceedings of the EXAG Workshop
  at AIIDE}, 2019.

\bibitem{sarkar2020exploring}
A.~Sarkar, A.~Summerville, S.~Snodgrass, G.~Bentley, and J.~Osborn, ``Exploring
  level blending across platformers via paths and affordances,'' in
  \emph{Proceedings of the AAAI Conference on Artificial Intelligence and
  Interactive Digital Entertainment}, vol.~16, no.~1, 2020, pp. 280--286.

\bibitem{khameneh2020embedding}
N.~Y. Khameneh and M.~Guzdial, ``Entity embedding as game representation,'' in
  \emph{Proceedings of the AIIDE Workshop on Experimental AI in Games}, 2020.

\bibitem{goodfellow2020generative}
I.~Goodfellow, J.~Pouget-Abadie, M.~Mirza, B.~Xu, D.~Warde-Farley, S.~Ozair,
  A.~Courville, and Y.~Bengio, ``Generative adversarial networks,''
  \emph{Communications of the ACM}, vol.~63, no.~11, pp. 139--144, 2020.

\bibitem{arjovsky2017wasserstein}
M.~Arjovsky, S.~Chintala, and L.~Bottou, ``Wasserstein generative adversarial
  networks,'' in \emph{International conference on machine learning}.\hskip 1em
  plus 0.5em minus 0.4em\relax PMLR, 2017, pp. 214--223.

\bibitem{Kingma2014}
D.~P. Kingma and M.~Welling, ``{Auto-Encoding Variational Bayes},'' in
  \emph{2nd International Conference on Learning Representations, {ICLR} 2014,
  Banff, AB, Canada, April 14-16, 2014, Conference Track Proceedings}, 2014.

\bibitem{larsen2016autoencoding}
A.~B.~L. Larsen, S.~K. S{\o}nderby, H.~Larochelle, and O.~Winther,
  ``Autoencoding beyond pixels using a learned similarity metric,'' in
  \emph{International conference on machine learning}.\hskip 1em plus 0.5em
  minus 0.4em\relax PMLR, 2016, pp. 1558--1566.

\bibitem{szegedy2015going}
C.~Szegedy, W.~Liu, Y.~Jia, P.~Sermanet, S.~Reed, D.~Anguelov, D.~Erhan,
  V.~Vanhoucke, and A.~Rabinovich, ``Going deeper with convolutions,'' in
  \emph{Proceedings of the IEEE conference on computer vision and pattern
  recognition}, 2015, pp. 1--9.

\bibitem{ioffe2015batch}
S.~Ioffe and C.~Szegedy, ``Batch normalization: Accelerating deep network
  training by reducing internal covariate shift,'' in \emph{International
  conference on machine learning}.\hskip 1em plus 0.5em minus 0.4em\relax pmlr,
  2015, pp. 448--456.

\bibitem{karras2018progressive}
\BIBentryALTinterwordspacing
T.~Karras, T.~Aila, S.~Laine, and J.~Lehtinen, ``Progressive growing of {GAN}s
  for improved quality, stability, and variation,'' in \emph{International
  Conference on Learning Representations}, 2018. [Online]. Available:
  \url{https://openreview.net/forum?id=Hk99zCeAb}
\BIBentrySTDinterwordspacing

\bibitem{gulrajani2017improved}
I.~Gulrajani, F.~Ahmed, M.~Arjovsky, V.~Dumoulin, and A.~C. Courville,
  ``Improved training of wasserstein gans,'' \emph{Advances in neural
  information processing systems}, vol.~30, 2017.

\bibitem{chollet2015keras}
F.~Chollet \emph{et~al.}, ``Keras,'' \url{https://keras.io}, 2015.

\bibitem{smith2009rhythm}
G.~Smith, M.~Treanor, J.~Whitehead, and M.~Mateas, ``Rhythm-based level
  generation for 2d platformers,'' in \emph{Proceedings of the 4th
  international Conference on Foundations of Digital Games}, 2009, pp.
  175--182.

\bibitem{horn2014comparative}
B.~Horn, S.~Dahlskog, N.~Shaker, G.~Smith, and J.~Togelius, ``A comparative
  evaluation of procedural level generators in the mario ai framework,'' in
  \emph{Foundations of Digital Games 2014, Ft. Lauderdale, Florida, USA
  (2014)}.\hskip 1em plus 0.5em minus 0.4em\relax Society for the Advancement
  of the Science of Digital Games, 2014, pp. 1--8.

\bibitem{marino2015empirical}
J.~Mari{\~n}o, W.~Reis, and L.~Lelis, ``An empirical evaluation of evaluation
  metrics of procedurally generated mario levels,'' in \emph{Proceedings of the
  AAAI Conference on Artificial Intelligence and Interactive Digital
  Entertainment}, vol.~11, no.~1, 2015, pp. 44--50.

\bibitem{canossa2015towards}
A.~Canossa and G.~Smith, ``Towards a procedural evaluation technique: Metrics
  for level design,'' in \emph{The 10th International Conference on the
  Foundations of Digital Games}.\hskip 1em plus 0.5em minus 0.4em\relax sn,
  2015, p.~8.

\bibitem{Summerville_2018}
\BIBentryALTinterwordspacing
A.~Summerville, ``Expanding expressive range: Evaluation methodologies for
  procedural content generation,'' \emph{Proceedings of the AAAI Conference on
  Artificial Intelligence and Interactive Digital Entertainment}, vol.~14,
  no.~1, pp. 116--122, Sep. 2018. [Online]. Available:
  \url{https://ojs.aaai.org/index.php/AIIDE/article/view/13012}
\BIBentrySTDinterwordspacing

\bibitem{sarkar2021generating}
A.~Sarkar and S.~Cooper, ``Generating and blending game levels via
  quality-diversity in the latent space of a variational autoencoder,'' in
  \emph{Proceedings of the 16th International Conference on the Foundations of
  Digital Games}, 2021, pp. 1--11.

\bibitem{sarkar2021dungeon}
------, ``Dungeon and platformer level blending and generation using
  conditional vaes,'' in \emph{2021 IEEE Conference on Games (CoG)}.\hskip 1em
  plus 0.5em minus 0.4em\relax IEEE, 2021, pp. 1--8.

\bibitem{rizzo2016energy}
M.~L. Rizzo and G.~J. Sz{\'e}kely, ``Energy distance,'' \emph{wiley
  interdisciplinary reviews: Computational statistics}, vol.~8, no.~1, pp.
  27--38, 2016.

\bibitem{feydy2019interpolating}
J.~Feydy, T.~S{\'e}journ{\'e}, F.-X. Vialard, S.-i. Amari, A.~Trouve, and
  G.~Peyr{\'e}, ``Interpolating between optimal transport and mmd using
  sinkhorn divergences,'' in \emph{The 22nd International Conference on
  Artificial Intelligence and Statistics}, 2019, pp. 2681--2690.

\bibitem{rosenblatt1956remarks}
M.~Rosenblatt, ``Remarks on some nonparametric estimates of a density
  function,'' \emph{The annals of mathematical statistics}, pp. 832--837, 1956.

\bibitem{mirza2014conditional}
M.~Mirza and S.~Osindero, ``Conditional generative adversarial nets,''
  \emph{arXiv preprint arXiv:1411.1784}, 2014.

\end{thebibliography}
\bibliographystyle{IEEEtran}
\end{document}